\documentclass[opre,sblanonrev]{informs4} % current default for manuscript submission

\OneAndAHalfSpacedXI

\usepackage{anyfontsize}
\DeclareMathVersion{bold}
\usepackage{color}
\usepackage{subfigure}
\usepackage{appendix}
\usepackage{amsmath, amsfonts}
\usepackage{bm}

\usepackage{accents}

\usepackage{algorithm2e}

\usepackage{natbib}
 \bibpunct[, ]{(}{)}{,}{a}{}{,}%
 \def\bibfont{\footnotesize}%

\usepackage{multirow, makecell}

\newcommand{\E}{{\rm E}}

\usepackage{natbib}
 \bibpunct[, ]{(}{)}{,}{a}{}{,}%
 \def\bibfont{\footnotesize}%

\theoremstyle{TH}
\newtheorem{property}{Property}
\theoremstyle{TH}
\newtheorem{condition}{Condition}
\theoremstyle{TH}

\usepackage[bookmarks=false, hypertexnames=false]{hyperref}
\hypersetup{
   colorlinks=true,
   linkcolor= [RGB]{17, 14, 178},% [RGB]{255,0,127}, %[RGB]{17, 14, 178}, %[RGB]{17, 14, 178},
   citecolor= [RGB]{17, 14, 178}, % [RGB]{255,0,127},
   filecolor= [RGB]{17, 14, 178},% [RGB]{255,0,127}, %[RGB]{17, 14, 178}, %[RGB]{17, 14, 178},      
   urlcolor= [RGB]{17, 14, 178},% [RGB]{255,0,127}, %[RGB]{17, 14, 178},
	pdfpagemode=FullScreen,
   }
   
\TheoremsNumberedThrough     % Preferred (Theorem 1, Lemma 1, Theorem 2)
\ECRepeatTheorems
\JOURNAL{Operations Research}

\EquationsNumberedThrough    % Default: (1), (2), ...
\MANUSCRIPTNO{}

\begin{document}
%\setcounter{page}{1}
%\DOI{10.1287/xxxx.0000.0000}%


\RUNTITLE{UCB for Large-Scale Pure Exploration: Beyond Sub-Gaussianity}

\TITLE{\large UCB for Large-Scale Pure Exploration: Beyond Sub-Gaussianity}

\RUNAUTHOR{Li, Fan, and Hong} % for blind review
% \ARTICLEAUTHORS{%
% \AUTHOR{First Author}
% %,\textsuperscript{a} Second Author,\textsuperscript{b} Third Author,\textsuperscript{c} Fourth Author,\textsuperscript{c}

% \AFF{INFORMS, 5521 Research Park Drive, Suite 200, Catonsville, Maryland 21228 \EMAIL{mirko.janc@informs.org}}
% %\textsuperscript{b}School of Industrial Engineering, Good College, Collegeville, Maine 01234 \EMAIL{secauth@goodcoll.edu}; 
% %\textsuperscript{c}Their Common Affiliation \EMAIL{thauth@anywhere.edu, fourauth@anywhere.edu}

% %mirko.janc@informs.org
% \AUTHOR{Second Author}

% \AFF{School of Industrial Engineering, Good College, Collegeville, Maine 01234, \EMAIL{secauth@goodcoll.edu}}

% \AUTHOR{Third Author, Fourth Author}

% \AFF{Their Common Affiliation \{thauth@anywhere.edu, fourauth@anywhere.edu\}}
% }
\ARTICLEAUTHORS{%
    % \AUTHOR{Zaile Li}
    % \AFF{School of Management, Fudan University, Shanghai, China, \\\EMAIL{zaileli21@m.fudan.edu.cn}}
    \AUTHOR{Zaile Li}
    \AFF{Department of Industrial Systems Engineering and Management, National University of Singapore, Singapore \\\EMAIL{zaile.li@nus.edu.sg}}
    
    \AUTHOR{Weiwei Fan}
    \AFF{Advanced Institute of Business and School of Economics and Management, Tongji University, Shanghai, China \\\EMAIL{wfan@tongji.edu.cn}} %, \URL{}}
    \AUTHOR{L. Jeff Hong}
    \AFF{Department of Industrial and Systems Engineering, University of Minnesota, Minneapolis, Minnesota\\\EMAIL{lhong@umn.edu}}
} % end of the block
\ABSTRACT{%
Selecting the best alternative from a finite set is the central objective of ranking and selection (R\&S) and best arm identification (BAI). Traditional R\&S or BAI approaches have predominantly relied on Gaussian or sub-Gaussian assumptions on the performance distributions of all alternatives, which limit their applicability to non-sub-Gaussian—especially heavy-tailed—problems. The need to move beyond sub-Gaussianity may become even more critical in large-scale problems, which tend to be especially sensitive to distributional specifications. In this paper, motivated by the widespread use of upper confidence bound (UCB) algorithms in sequential decision making, we investigate their performance in large-scale, non-sub-Gaussian R\&S settings. We consider the simplest category of UCB algorithms, where the UCB value for each alternative is defined as the sample mean plus an exploration bonus that depends only on its own sample size. 
We abstract this into a meta-UCB algorithm and propose letting it select the alternative with the largest sample size as the best upon stopping. 
For this meta-UCB algorithm, we first derive a distribution-free lower bound on the probability of correct selection. 
Building on this bound, we first study the indifference-zone formulation and show that the meta-UCB algorithm—and therefore a broad class of UCB algorithms—achieves sample optimality as long as the \emph{variances are uniformly bounded}. We then extend sample optimality to a non-indifference-zone configuration with polynomially shrinking mean gaps under the same variance assumption.
% , using a variance-calibrated exploration bonus of order $\log(n)/\sqrt n$. The admissible gap-shrinkage range is the full range compatible with a linear aggregate inverse-gap cost. 
These results demonstrate the applicability of UCB algorithms to large-scale R\&S problems with non-sub-Gaussian distributions. Numerical experiments support our results and provide additional insights into the comparative behaviors of UCB algorithms within and beyond our meta-UCB framework.

% The abstract is limited to one paragraph and should contain no references and 
% no equations. Following the abstract, please enter the following items 
% (depending on the requirements of the particular INFORMS journal): (1) key 
% words ({KEYWORDS}), (2) MSC subject classification identifying primary and 
% secondary codes (see http://www.ams.org/msc) ({MSCCLASS}), (3) OR/MS 
% classification, also identifying primary and secondary (see 
% http://or.pubs.informs.org/Media/ORSubject.pdf) ({ORMSCCLASS}), (4) subject 
% classifications ({SUBJECTCLASS}), and (5) area of review ({AREAOFREVIEW}). In 
% later stages of manuscript processing, the history line ({HISTORY}) will be 
% added.
}
% \AREAOFREVIEW{Simulation}

\maketitle
% \section{The Greedy Algorithm}
\section{Introduction}

% Our requirements are inspired by recent boundary-crossing ideas, and indicate how the confidence band impact the probability of correct selection for the UCB

% , as the exploitation occurs only once following the exploration process

Ranking and selection (R\&S) concerns identifying the best alternative—the one with the highest mean performance—from a finite set. The true means of the alternatives are unknown, but the decision-maker may collect noisy observations through stochastic simulation or experimentation to estimate them and make an informed selection. This best-seeking problem under uncertainty has been extensively studied in two parallel literatures: as R\&S in stochastic simulation \citep{bechhofer1954single, kim2006selecting}, and as best arm identification (BAI) within the pure-exploration literature in machine learning \citep{audibert2010best}. Despite using different terminology, both frameworks share the same fundamental goal: to design efficient algorithms that adaptively allocate sampling effort across the alternatives, maximize selection accuracy, and minimize the total sampling cost.

A wide range of algorithmic approaches have been developed for efficient R\&S. Prevalent examples include pairwise comparison \citep{kim2006selecting}, optimal computing budget allocation \citep{glynn2004large, chen2011stochastic}, Bayesian expected value of information \citep{frazier2008knowledge}, sequential elimination \citep{karnin2013almost}, and Thompson sampling \citep{russo2020simple, qin2025dual}, among others. We refer readers to \citet{hong2021review} and \citet{lattimore2020bandit} for comprehensive reviews. Among the many approaches, one particularly interesting idea is the use of upper confidence bound (UCB) algorithms \citep{lai1987, auer2002finite}—a cornerstone of the multi-armed bandit literature. While UCB was originally designed for cumulative reward maximization in online learning, its structural simplicity—making sampling decisions based solely on per-alternative UCB indices—has made it widely applicable to many learning tasks (see, e.g., \citealt{agrawal2019recent}), with BAI being no exception (see, e.g., \citealt{audibert2010best} and \citealt{jamieson2014best}). These works have made significant theoretical progress and have found important applications in areas such as Monte Carlo tree search \citep{fu2016alphago}, drug discovery \citep{terayama2018machine}, and prompt engineering for large language models \citep{pryzant2023automatic}.

A predominant assumption in these R\&S and BAI works is that the performance distributions of alternatives are Gaussian or, more generally, sub-Gaussian. While this assumption offers significant analytical convenience and often forms the foundation for algorithm design, it also limits the scope of applicability. In R\&S, assuming Gaussianity is a longstanding convention, typically justified by central limit theorem arguments—the batch mean of simulation outputs is often approximately normal \citep{kim2006selecting}. This normality assumption supports a range of analytical tools, such as Brownian motion approximations and Bayesian conjugacy (see, e.g., \citealt{fan2016indifference} and \citealt{chick2001new}). In BAI, the assumption of sub-Gaussianity is more prevalent. Sub-Gaussian distributions include many commonly encountered families, such as Gaussian, Bernoulli, and all bounded distributions, making them broadly accepted in the design of BAI algorithms (see, e.g., \citealt{jamieson2014lil}). 

However, non-sub-Gaussian problems frequently arise in practice. In light-tailed problems, the alternative distributions exhibit exponential tail decay but are slower than sub-Gaussian ones; the exponential distribution is a canonical example, commonly seen in the analysis of queueing systems \citep{gao2016optimal, zhang2020sequential}. In heavy-tailed problems, the distributions have tails that decay more slowly than exponentially, for example, sub-exponentially in the lognormal family, or polynomially in the Pareto family. Heavy-tailed distributions are particularly relevant in contexts where extreme values are more likely, including reliability engineering \citep{john2006lognormal, chen2011enhanced}, healthcare operations \citep{strum2000modeling}, financial portfolio management \citep{panahi2016model}, and network control \citep{abry2010revisiting}.
Non-sub-Gaussianity introduces significant challenges for R\&S, as many tools and theoretical guarantees built on sub-Gaussianity are no longer valid.
These challenges have recently motivated growing interest in developing methodologies for handling general, especially heavy-tailed, distributions (see, e.g., \citealt{glynn2018selecting} and \citealt{agrawal2020optimal}). A detailed review of recent progress is provided in Section \ref{subsec: literature}.
% \citealt{lee2016general}, 
% progress remains in its early stages.

% \textit{For UCB, there has also been a broad interest in their performance in heavy-tailed settings.}

% true distribution type of alternative performances may be unknown to the decision-maker prior to conducting experiments. In such cases, 

% , regardless of the actual distribution class
The importance of carefully addressing non-sub-Gaussianity is particularly pronounced when the number of alternatives $k$ is large. In practice, the true distributional form of each alternative is often unknown, and a pragmatic strategy is to apply prominent sub-Gaussian algorithms heuristically. Interestingly, despite the lack of theoretical guarantees, this approach has sometimes been shown to perform well in small-scale problems, where $k$ is relatively small (e.g., fewer than 20), even under heavy-tailed distributions \citep{nelson2001simple, shin2018tractable}. One intuition may come from the \textit{high-confidence asymptotic regime}, where $k$ is fixed and the probability of correct selection (PCS) is driven close to one \citep{dong2016three}. In this regime, the total sampling budget $B$ grows large, and each alternative is allocated a growing number of observations. As $B$ increases, the sample averages converge, and by the central limit theorem, they may approximate normality. However, as the number of alternatives $k$ increases, the reliability of this strategy deteriorates. \citet{nelson2001simple} and \citet{shin2018tractable} demonstrate that Gaussian-specific algorithms can suffer substantial performance loss under heavy-tailed distributions when $k$ grows large (e.g., $k \approx 500$). This degradation likely stems from the accumulation of misspecification errors introduced by extreme values across alternatives—errors that may be negligible in small-scale problems but become significantly amplified in large-scale ones. These observations highlight the necessity of designing and analyzing algorithms that remain effective in large-scale, non-sub-Gaussian environments.
Coinciding with the challenges posed by non-sub-Gaussianity, recent years have seen a growing need to address large-scale R\&S problems involving a significant number of alternatives \citep{jamieson2013finding, luo2015fully, ni2017efficient}. Practical examples, such as those studied by \cite{luo2015fully} and \cite{pei2022parallel}, feature thousands to over one million alternatives. To model these settings,  \cite{jamieson2013finding} and later \cite{zhong2022knockout} consider a new \textit{large-scale asymptotic regime}, in which $k \to \infty$. Under this regime, understanding how an algorithm’s required sampling budget scales with \(k\) becomes crucial. Sample-optimal algorithms—those requiring a budget growing at the minimal possible rate, often \(\mathcal{O}(k)\)—are particularly desirable and have become a key benchmark for large-scale R\&S \citep{hong2022solving, zhang2024sample, li2023surprising}. However, this large-scale regime challenges the assumptions that underlie small-scale R\&S algorithms. In particular, the central limit theorem argument, which assumes a sufficiently large number of observations per alternative, may no longer hold. With only $\mathcal{O}(k)$ total sampling budget, the average number of observations per alternative remains bounded, and the cumulative effect of increased likelihood of extreme values can become severe as $k$ grows.
% These findings further underscore the importance of explicitly addressing non-sub-Gaussianity in large-scale pure exploration.
Yet, to the best of our knowledge, existing works in this regime continue to assume Gaussianity or sub-Gaussianity, leaving this critical issue of non-sub-Gaussianity largely unexplored.

In this work, motivated by the versatility of UCB algorithms, we explore their performance in the challenging setting of large-scale R\&S with non-sub-Gaussian distributions. We find that UCB algorithms achieve sample optimality in this regime as long as the variances of all alternatives are uniformly bounded. We study a broad class of UCB algorithms, develop a unified analysis of their PCS, and establish sufficient conditions under which they achieve sample optimality. These findings may offer a promising approach to addressing the challenge of moving beyond sub-Gaussianity in large-scale R\&S and further highlight the applicability of UCB algorithms in R\&S.
% Specifically, we demonstrate that under a characterization of UCB algorithms, they can maintain an asymptotically non-zero PCS with a total sampling budget \(T = \mathcal{O}(k)\) as \(k \to \infty\), even under general distributional assumptions.

 % or a large number of alternatives \citep{ bayati2020unreasonable}

% In this paper, we take the first step of moving beyond sub-Gaussianity in solving large-scale pure exploration problems. Towards this goal, we start by analyzing the performance of the UCB algorithms in solving non-sub-Gaussian pure exploration problems because of their enduring popularity and widespread applications \citep{tanczos2017kl, pryzant2023automatic, zhang2024language}, which also responds to the growing interest in understanding their behaviors in unconventional settings like those with heavy-tailed observations \citep{bubeck2013bandits} or a large number of alternatives \citep{ bayati2020unreasonable}. With this focus, surprisingly, we find that for a broad class of UCB algorithms, they may achieve the sample optimality in solving large-scale pure exploration problems in the absence of sub-Gaussianty. 

% we consider UCB algorithms where the index value of each alternative depends only on its own prior information (if any) and its sample history. This class represents a major category of UCB algorithms. More specifically,
UCB algorithms make sampling decisions in a sequential manner, typically selecting at each round the alternative with the highest current UCB value. The definition of the UCB value varies across algorithmic variants. In this work, we focus on a class of UCB algorithms in which the UCB value for each alternative is defined as the sample mean plus a non-negative bonus term that depends deterministically on its own sample size. This class represents the simplest category of UCB algorithms. 
To support a unified analytical framework, we consider a meta-UCB algorithm in which the bonus function is arbitrary but required to vanish asymptotically—that is, it converges to zero as the sample size approaches infinity. This meta-algorithm encompasses a number of well-known UCB variants, including UCBE \citep{audibert2010best} and MOSS \citep{audibert2009minimax}. Additional examples are discussed in Section \ref{subsec: meta-UCB}. We adopt a fixed-budget formulation of R\&S, in which the total sampling budget is predetermined. Under this formulation, a decision rule (or selection standard) is applied at the end of the sampling process to select one alternative as the best, based on the collected sample information. While the most common rule is to select the alternative with the largest sample mean, we consider an alternative rule that selects the alternative with the \emph{largest sample size} \citep{jamieson2014lil, andradottir2009balanced}. Interestingly, this selection rule may facilitate a simplified analysis and enable us to obtain a rich set of theoretical results.
% We also discuss and evaluate alternative selection criteria in Section \ref{subsec: num_selection}.

% —it does not rely on any assumptions about the distributions of the alternatives
We begin our analysis of the meta-UCB algorithm by studying its PCS. Despite the clear structure of UCB algorithms, analyzing their PCS is far from straightforward due to their sequential and adaptive nature. Existing analyses often rely heavily on sub-Gaussian assumptions and result in vacuous (e.g., negative) PCS lower bounds when applied to large-scale problems. Inspired by ideas from \citet{li2023surprising}, we approach the analysis from a boundary-crossing perspective. Specifically, we treat the stochastic UCB process (defined in Section \ref{subsec: boundary-crossing}) of the best alternative as a boundary and view the UCB processes of the non-best alternatives as crossing below this boundary during the sampling process. This perspective enables us to derive upper bounds on the number of observations allocated to each non-best alternative in terms of their boundary-crossing times. Subsequently, the selection standard of the largest sample size enables us to establish a clean PCS lower bound for the meta-UCB algorithm. Notably, this PCS lower bound is distribution-free. Moreover, for any specific UCB algorithm within the meta class, its PCS lower bound can be obtained directly by substituting its bonus function into the lower bound. The PCS lower bound also provides intuitive structural insights into the dual role of the bonus function in UCB algorithms: enabling effective exploration while controlling the associated sampling cost. See Section \ref{subsec: insight} for a more detailed discussion.
% A more aggressive (e.g., wider) bonus function increases the likelihood that the true best alternative is sampled more often and ultimately selected—but at the expense of amplifying the boundary-crossing times of the non-best alternatives, thus increasing overall sampling cost—and vice versa.

Building on the PCS analysis, we examine the performance of the meta-UCB algorithm in solving large-scale, non-sub-Gaussian R\&S problems. We begin with the indifference-zone (IZ) formulation, which assumes that the mean gap between the best and second-best alternatives remains constant as the number of alternatives increases. This formulation, also referred to as the sparse configuration \citep{jamieson2013finding}, is standard in the study of sample optimality \citep{hong2022solving, li2023surprising}. Under the IZ formulation and uniformly bounded variances, we prove that the meta-UCB algorithm achieves sample optimality: with a total sampling budget that grows linearly with $k$, the algorithm maintains a non-zero PCS asymptotically.
% open new opportunities for designing efficient large-scale pure exploration algorithms.

% The sample optimality of UCB algorithms in large-scale pure exploration settings has remained unknown, and their robustness to non-sub-Gaussianty is unexpected. These findings underscore the versatility of the UCB approach in pure exploration, extending its applicability beyond the traditional small-scale setting. Besides, it is also interesting to note that some of the included UCB algorithms are designed for MAB problems, which highlights a new linkage between pure exploration and MAB.

% The sample optimality of UCB algorithms in solving large-scale pure exploration problems has remained unknown, and their robustness to non-sub-Gaussianity is unexpected. These results confirm the versatility of the UCB approach beyond the traditional small-scale, sub-Gaussian pure exploration setting and open new opportunities for designing efficient large-scale pure exploration algorithms. We will discuss this direction in Section \ref{sec: conclude}. Besides, it is noteworthy that some of the UCB algorithms encompassed by the meta algorithm are originally designed for MAB problems, and their sample optimality highlights a new linkage between pure exploration and MAB. 

% Although the IZ formulation is standard, it is sometimes argued to be impractical for large-scale problems \citep{hunter2017parallel}. 
We then explore whether the meta-UCB algorithm can achieve the sample optimality beyond the IZ formulation, to further test the robustness of the UCB approach. Without the IZ assumption, the mean gap between the best and some other alternatives may become arbitrarily small as $k$ increases. In this challenging setting, rather than relaxing the objective to identifying a ``good enough'' alternative \citep{ni2017efficient, eckman2018guarantees}, we continue to focus on the PCS. To achieve the sample optimality, we identify a key condition: the UCB value of the best alternative should always exceed its true mean with non-zero probability. This requirement is natural for UCB algorithms—it aligns with the interpretation of anytime-valid confidence bounds—but may not be satisfied by existing IZ-requiring algorithms \citep{li2023surprising}. For a representative configuration with polynomially shrinking gaps, we construct such a confidence sequence. 
% The resulting bonus is $A\log_2(2n)/\sqrt n$ and is calibrated by a known uniform variance bound. 
Combining it with first- and second-moment controls for the boundary-crossing times proves sample optimality again under uniformly bounded variances whenever the gap exponent $\beta$ satisfies $\beta<1/2$. This positive result further highlights the robustness of UCB algorithms for large-scale R\&S.

Lastly, we conduct a series of numerical experiments to examine our theoretical findings and the qualitative role of exploration. Under the IZ formulation, the tested meta-UCB algorithms exhibit sample-optimal behavior with non-sub-Gaussian alternatives. We also examine the classic UCB1 algorithm of \citet{auer2002finite}, a representative example outside our meta-UCB class. It performs significantly worse and appears to lack sample optimality. To understand this performance gap, we analyze the budget-allocation behavior of the algorithms. Algorithms in the meta-UCB class display the predicted boundary-crossing dynamics, whereas UCB1 tends to over-explore by allocating the budget more uniformly across alternatives. This comparison highlights the importance of our selected class of UCB algorithms.

% Additionally, we also examine UCB variants that fall outside the scope of our meta-UCB algorithm. The outcomes in these cases depend on the specific structure of the UCB value and offer further insight into the behavior of different UCB algorithms in solving large-scale pure exploration problems; see Section~\ref{subsec: num_more_general} for a detailed discussion.

It is important to acknowledge that our perspective in this paper is primarily theoretical. We do not aim to design new UCB algorithms, or optimize the choice of bonus functions, or address practical implementation issues. That said, we believe the proposed new results and insights can serve as a foundation for future work aimed at designing practically effective UCB algorithms for large-scale, non-sub-Gaussian (especially heavy-tailed) R\&S problems. In addition, we emphasize that our analysis focuses on algorithms with UCB values defined as the sample mean plus a bonus function. Other forms of UCB are also worth exploring, particularly those based on robust mean estimators—such as truncated or weighted sample means—that are designed to handle heavy-tailed distributions \citep{bubeck2013bandits, glynn2018selecting}. We discuss and conjecture in Section \ref{subsec: more_general} that the sample optimality results may extend to these broader classes when the UCB values of different alternatives are decoupled—that is, each value depends only on the alternative’s own sample size and observations.  However, formally verifying these results would likely require case-by-case analysis, which we leave for future work.

% Numerical results in Section \ref{subsec: num_more_general} offer some support for this conjecture. 
% Our analysis may serve as a template for analyzing such more general cases.  

% It is also important to note that our sample optimality results rely on the assumption of a deterministic bonus function dependent solely on the sample size of each alternative. While this assumption limits the scope of our framework, it provides a tractable structure for boundary-crossing analysis. Interesting, numerical studies reveal that UCB algorithms beyond this meta-framework may fail to achieve sample optimality. We will explore this phenomenon in Section \ref{subsec: num_other}.

% Alongside the sample optimality, we also analyze the non-asymptotic  problem-dependent sample complexity of specific UCB algorithms to better characterize their performance in solving pure exploration problems. We consider two specific bonus functions such that the PCS lower bound is no smaller than a target level for sub-Gaussian and non-sub-Gaussian problems, respectively. For each bonus function, we then derive the sample complexity of the correspondent UCB algorithm. The results also display the sample optimality of the UCB algorithms. Interestingly, the sample complexity for the sub-Gaussian case matches a known lower bound for sub-Gaussian problems, indicating the effectiveness of our boundary-crossing analysis.

\subsection{Related Literature}
\label{subsec: literature}

The R\&S and BAI literatures have proposed several approaches to relax the (sub-)Gaussianity assumption. The central-limit-theorem approach approximates the distribution of scaled partial sums of non-Gaussian observations with a Gaussian limit in asymptotic regimes where the sample size of each alternative tends to infinity \citep{toscano2015asymptotic, lee2016general, fan2016indifference}. The large-deviation (LD) approach, pioneered by \cite{glynn2004large}, leverages the LD rate function of each alternative to solve the optimal budget allocation problem and is naturally suited for light-tailed observations (see, e.g., \citealt{gao2017new}). It has also been extended by \cite{broadie2007implications} and \cite{blanchet2008large} to accommodate heavy-tailed distributions; see \cite{dong2016three} for an overview of these two approaches. A third class of methods employs robust mean estimators—such as the truncated sample mean—to mitigate the influence of heavy tails \citep{yu2018pure, glynn2018selecting}. 
Most recently, \citet{agrawal2020optimal, agrawal2021optimal} introduce a change-of-measure technique for minimizing the sampling budget needed to achieve a target PCS in heavy-tailed settings. In addition, there are also algorithmic approaches, including those of \cite{garivier2016optimal} and \cite{russo2020simple}, that naturally accommodate alternative distributions from the same one-parameter exponential family.
While these approaches provide valuable tools and insights for handling non-sub-Gaussian distributions, they are primarily developed for small-scale problems. In contrast, our work focuses on the large-scale setting and demonstrates that simple UCB algorithms can remain robust and sample-efficient even in the presence of heavy tails.
% , shin2018tractable, chen2022balancing

% chan2019optimal,lee2022minimax
Our work also responds to the growing interest in understanding the fundamental behaviors of UCB algorithms (see, e.g., \citealt{fan2024fragility}) and their performance in unconventional settings, such as learning with heavy-tailed observations \citep{bubeck2013bandits} or with a large number of arms \citep{chan2019optimal, bayati2020unreasonable}. These studies often report negative results, showing that traditional UCB algorithms may suffer from sub-optimal performance in such challenging environments. In contrast, we present sample optimality results for large-scale R\&S under mild distributional assumptions. These results may offer new insights into their performance and further highlight their potential for R\&S.
% , a broad class of UCB algorithms can still achieve sample optimality
% It is also noteworthy that several UCB algorithms encompassed by our meta framework are known to be optimal in classical MAB problems. Their sample optimality in pure exploration thus highlights a new and intriguing connection between multi-armed bandits and best-arm identification.

The remainder of this paper is organized as follows. In Section \ref{sec: problem}, we provide the problem formulation and introduce the meta-UCB algorithm. In Section \ref{sec: PCSanalysis}, we analyze the PCS of the meta-UCB algorithm and examine the role of the bonus function. In Section \ref{sec: sampleoptimality}, we establish the sample optimality of the meta-UCB algorithm under the IZ formulation. In Section \ref{sec: extensions}, we explore extensions beyond the IZ formulation and discuss broader classes of UCB algorithms not covered by our analysis.
In Section \ref{sec: numerical}, we conduct numerical experiments to verify our theoretical results. Finally, we conclude the paper in Section \ref{sec: conclude} and include supplementary technical details in the E-Companion.

\section{Problem Statement}
\label{sec: problem}
% \vspace{6pt}
In this section, we first introduce the notation and preliminaries for large-scale R\&S. We then present a general class of upper confidence bound (UCB) algorithms, referred to as the meta-UCB algorithm, which serves as our primary focus for addressing this problem under general distributional settings.

\subsection{Notations and Preliminaries}
\label{subsec: notations}
Let $k$ denote the number of alternatives in an R\&S problem, which represents the scale of the problem. For each alternative $i = 1, \dots, k$, let $X_i$ be a random variable representing its performance, and let $\mu_i = \E[X_i]$ denote its unknown mean. The objective is to identify the best alternative—that is, the one with the largest mean performance $\max_{i=1, \cdots, k} \mu_i$. Without loss of generality, we assume that alternative 1 is the best. Since the means $\mu_i$ are unknown, one must sample the alternatives—i.e., collect independent and identically distributed (i.i.d.) observations $X_{i,j}$, $j = 1, 2, \dots$, from each alternative $i$—to obtain sample information, so that an informed selection decision can be made. We assume throughout that the observation streams of distinct alternatives are mutually independent. In this paper, we adopt the fixed-budget formulation, where the total sampling budget $B$ is predetermined. By sequentially allocating this budget across alternatives, an algorithm selects one alternative, denoted by $i^*$,  as the best based on the collected sample information, once the budget is exhausted. The efficiency of such an algorithm is measured by the probability of correct selection (PCS), defined as $\Pr \{i^* = 1\}$.
% the probability that the algorithm correctly selects the best alternative

The performance of R\&S algorithms depends heavily on the distributional characteristics of the alternatives. Conventionally, it is often assumed that each $X_i$ follows a Gaussian or, more generally, a sub-Gaussian distribution \citep{kim2006selecting, audibert2010best}. While these assumptions offer technical convenience, they may limit the applicability of the resulting algorithms. In this paper, we explore settings where such assumptions do not hold—specifically, non-sub-Gaussian problems, including heavy-tailed cases. 
The challenges posed by non-sub-Gaussianity may become particularly pronounced in large-scale problems, where the number of alternatives $k$ is large.  In large-scale R\&S, the key performance measure is how the total sampling budget must grow with $k$ to maintain meaningful selection performance, specifically, a non-zero PCS. This scaling behavior has been formalized in recent work \citep{jamieson2013finding, tanczos2017kl, zhong2022knockout}, which shows that achieving a non-zero PCS as $k \to \infty$ requires a total sampling budget that grows at least at order $\Omega(k)$. An algorithm that meets this lower bound can be said to be \emph{sample optimal}. In the fixed-budget setting, \citet{hong2022solving} formalize this idea and define the sample optimality as follows.
 % 
% by adopting an asymptotic regime in which $k \rightarrow \infty$, 
\vspace{0.2cm}
\begin{definition}[Sample Optimality]
    \label{def: rate_optimality_PCS}
    An R\&S algorithm is sample optimal if there exists a pair of constants $\alpha \in (0, 1)$ and $c > 0$ such that
    \begin{equation}\label{eqn:ro}
        \liminf_{k\to\infty}\, \mathrm{PCS} >\alpha\ {\rm for}\ B=ck.
    \end{equation}
\end{definition}
\vspace{0.2cm}

% Distribution-free should not be mentioned here.
% Importantly, the definition is distribution-free, meaning that it does not rely on any specific assumptions about the underlying distributions of the alternatives. 
% This notion of sample optimality provides a natural benchmark for measuring the effectiveness of algorithms in large-scale pure exploration. Despite this generality, existing research on large-scale pure exploration has focused exclusively on sub-Gaussian problems. In this paper, motivated by the widespread popularity of Upper Confidence Bound (UCB) algorithms across various learning-based decision-making tasks—as well as their conceptual simplicity and empirical success—we investigate whether UCB algorithms can address this challenge.
This notion of sample optimality provides a benchmark for evaluating algorithm performance in large-scale R\&S and serves as the primary metric in this paper. While existing work on large-scale R\&S has focused exclusively on problems where all alternatives follow sub-Gaussian distributions, we investigate the performance of UCB algorithms in the more challenging non-sub-Gaussian regime.

\subsection{A Meta-UCB Algorithm}
\label{subsec: meta-UCB}

% Due to their simple index-based structure, UCB algorithms have been extensively studied and applied across various sequential decision-making problems, with pure exploration being no exception. 
UCB algorithms allocate the sampling budget sequentially—i.e., one observation at a time—based on a UCB value assigned to each alternative. This value serves as an ``optimistic'' estimate of the unknown mean and is typically defined as the upper bound of a confidence interval. The gap between the UCB value and the sample mean captures the \emph{exploration bonus} \citep{lattimore2020bandit}. A wide variety of UCB algorithms have been proposed, differing in how the UCB values are constructed. In this work, we focus on a class of simple UCB algorithms in which the UCB value for each alternative $i$ is defined as the sum of the sample mean $\bar X_i(n_i) = \frac{1}{n_i} \sum_{j=1}^{n_i} X_{i,j}$ and an exploration bonus function $f(n_i)$ that depends only on its sample size $n_i$. This setting represents a major category of UCB algorithms in which the UCB values are \textit{decoupled}—each depends only on the alternative's own sample information, i.e., its sample size and observations. To unify this class, we introduce a meta-UCB algorithm that allows for a general exploration bonus function. The algorithm is presented in Algorithm~\ref{algo: metaUCB}. For the exploration bonus function, we only require that it is non-negative and converges to zero as the sample size increases to infinity. This structure is formalized in the following assumption.
% prior information (if any) and its
% There are many different UCB algorithms with different definitions of UCB indexes in the literature. To unify the analysis, in this paper, we consider a meta-UCB algorithm with a generic UCB value which is detailed in Algorithm \ref{algo: metaUCB}. In Algorithm \ref{algo: metaUCB}, for each alternative $i=1, \cdots, k$, we use $n_i$ to keep track of its sample size  and let $\bar X_i(n_i)$ denote the sample mean of the first $n_i$ observations, i.e., $\bar X_i(n_i) = \frac{1}{n_i} \sum_{j=1}^{n_i} X_{i,j}$. 

% After the total sampling budget $B$ is exhausted (\emph{stopping rule}),  the algorithm returns the alternative with the largest sample size as the best alternative (\emph{implementation rule}). 

% At an initial stage, the algorithm takes one observation from each alternative to initialize the sample size and sample mean. Then, in the following stage, it keeps sampling the alternative with the largest UCB index. 

\RestyleAlgo{ruled}
\LinesNumbered
\SetAlgorithmName{Algorithm}{Algorithm}{Algorithm}
% \SetAlgoCaptionLayout{centerline}
\begin{algorithm}[hbtp]
\label{algo: metaUCB}
% \caption{() \textbf{Explore-First Top-$\mathbf{m}$ Greedy (EFGm) Algorithm}}
\caption{\textbf{Meta-UCB}}
    \label{algorithm: topM}
\KwIn{{$k$ alternatives $X_1,\ldots,X_k$}, the total sampling budget $B$}
\For{$i=1$ \KwTo $k$ }{
    take one observation $x_{i}$ from alternative $X_i$, set $n_i=1$ and initialize $U_i(n_i)$;
}
\While{$\sum_{i=1}^k n_i < B$ }{
\vspace{0.2cm}
select the alternative $s = { \argmax}_{i \in \{1, \ldots, k\}} U_i(n_i)$;

take one observation $x_{s}$ from $X_s$, set $n_s = n_s+1$, and update $U_s(n_s)$\;
}
\KwOut{alternative $b = { \argmax}_{i \in \{1, \ldots, k\}} n_i$}
\end{algorithm}

\begin{assumption}
\label{assu: bonusfunction}
For each alternative $i=1, \cdots, k$, the UCB value takes the form
$U_i(n_i) = \bar X_i(n_i) + f(n_i)$, where the exploration bonus function $f$ is non-negative and satisfies $\lim_{n_i\to\infty} f(n_i) = 0$.
\end{assumption}

% \begin{assumption}
%     For each alternative $i$, after $n_i$ observations have been collected, the UCB value $U_i(n_i)$ is a deterministic function of the prior information about $X_i$ (if any) and the observed samples $\{X_{i1}, X_{i2}, \ldots, X_{i,n_i}\}$. Furthermore, the index is asymptotically consistent, i.e., 
% $$
% U_i(n_i) \xrightarrow{a.s.} \mu_i \quad \text{as } n_i \to \infty.
% $$
% \end{assumption}

% \begin{remark}
%     In the meta-UCB algorithm, we use one observation to initialize the UCB index. Some UCB algorithms may require $n_0 > 1$ observations, to initialize the sample estimates of for statistics like variance. Simple modifications are suffice for our analysis to incorporate these cases. We maintain $n_0=1$ for clarity.  
% \end{remark}

Intuitively, this assumption requires that as the sample size $n_i \to \infty$, the exploration bonus associated with the sample estimate of the true mean $\mu_i$ vanishes, so that the UCB value $U_i(n_i)$ converges to $\mu_i$. This behavior is essential to ensure that the best alternative can be identified whenever all the alternatives have been sufficiently evaluated.
%This behavior aligns with the requirement of asymptotic consistency. 
The assumption is broad enough to encompass many well-known UCB algorithms. Three illustrative examples are provided below. Additional examples of bonus functions satisfying Assumption~\ref{assu: bonusfunction} can be found in, e.g., \citet{degenne2019bridging}, \citet{lattimore2020bandit}, \citet{zhong2021achieving}, and \citet{zhang2024fast}. 
% , especially in heavy-tailed settings \citep{bubeck2013bandits,lee2020optimal}
% Beyond this assumption, there exist other variants of particular interest that has the decoupling structure. These extensions 

% \vspace{0.2cm}
\begin{example}[The UCB-E Algorithm of \citealt{audibert2010best}]
\label{example: ucbe}
    $$U_i(n_i)  =  \bar X_i(n_i)  + \sqrt{\frac{a}{n_i}},$$
    where $a$ is a constant irrelevant to $n_i$.
    % \in \left(0, \frac{25}{36}\frac{B-k}{H_1}\right]$ and $H_1 = \sum_{i=2}^k \frac{1}{(\mu_i - \mu_1)^2}$. 
    % Under the IZ assumption (Assumption \ref{assu: iz}), $H_1$ may be viewed as $k/\gamma$ and $a$ may be chosen to be $\lambda c$.
\end{example}

% audibert2009minimax,
\begin{example}[The MOSS Algorithm of \citealt{audibert2010regret}]
\label{example: moss}
    $$U_i (n_i) = \bar X_i(n_i) + \sqrt{\frac{1}{n_i}  \max\left\{\log\left(\frac{c}{n_i}\right), 0\right\}},$$
where $c=B/k$.
\end{example}

% \begin{example}[The UCB($\delta$) Algorithm of \citealt{lattimore2020bandit}]
%     $$f(n_i) = \sqrt{\frac{2 \log(1/\delta)}{n_i}},$$
%     where $\delta$ is a constant irrelevant to $n_i$. 
% \end{example}

\begin{example}[The LiL-UCB Algorithm of \citealt{jamieson2014lil}]
\label{example: lil}
    $$U_i (n_i) = \bar X_i(n_i)  + (1+\beta)(1+\sqrt{\varepsilon}) \sqrt{\frac{2 \sigma^2(1+\varepsilon) \log \left(\frac{\log \left((1+\varepsilon) n_i\right)}{\delta}\right)}{n_i}},$$
where $\beta,\varepsilon, \sigma$ and $\delta$ are constants irrelevant to $n_i$. 
\end{example}
% \vspace{0.2cm}

%Admittedly, Assumption~\ref{assu: bonusfunction} describes only a subclass of UCB algorithms. Beyond this assumption, there exist other variants of particular interest, especially in heavy-tailed settings \citep{bubeck2013bandits,lee2020optimal}. These extensions are discussed further in Section~\ref{subsec: more_general}.

% Broadly speaking, UCB algorithms can be grouped into two categories. The first category—within which Assumption~\ref{assu: bonusfunction} falls—includes algorithms where the UCB value of each alternative is determined solely by its own prior and sample information. Beyond this assumption, there exist other variants of particular interest, especially in heavy-tailed settings. These extensions are discussed further in Section~\ref{subsec: more_general}. The second category includes algorithms where the index may also depend on global statistics, such as the total sample size of all alternatives. A prominent example in this class is the classical UCB1 algorithm of \cite{auer2002finite}. In Section~\ref{sec: numerical}, we examine the performance of UCB1 and find that, interestingly, it can behave quite differently—and may perform significantly worse—compared to algorithms satisfying Assumption~\ref{assu: bonusfunction}. For these reasons, we focus on the first category of UCB algorithms.
% \subsection{The Selection Rule}
% This introduces coupling among alternatives and may significantly alter algorithmic behavior. 

It is worth highlighting that the decoupling structure in UCB algorithms satisfying Assumption~\ref{assu: bonusfunction} has important implications for both their performance in R\&S and the techniques used to analyze them. We elaborate on these implications in Section~\ref{subsec: boundary-crossing} and Section~\ref{subsec: budget_allocation}. Readers familiar with the machine learning literature may recognize that many UCB algorithms,  such as the well-known {UCB1}  \citep{auer2002finite}, do not satisfy this decoupling structure. Interestingly, numerical experiments in Section~\ref{sec: numerical} show that {UCB1} behaves quite differently and may perform significantly worse compared to algorithms conforming to Assumption~\ref{assu: bonusfunction}. We will discuss algorithms such as {UCB1} that do not satisfy Assumption \ref{assu: bonusfunction} in Section~\ref{subsec: more_general}.

We would also like to highlight that in Algorithm~\ref{algo: metaUCB}, the alternative with the largest sample size is selected as the best when the total sampling budget is exhausted. This selection criterion, though less common, has been adopted in prior work—for example, by \citet{jamieson2014lil}—and, interestingly, has also appeared in early simulation optimization literature (see, e.g., \citealt{andradottir2009balanced}). While the more conventional and intuitive selection criterion is the largest sample mean, we will show that using the largest sample size as the selection criterion offers analytical advantages that facilitate performance analysis.

\section{PCS Analysis of the Meta-UCB Algorithm}
\label{sec: PCSanalysis}
In this section, we analyze the PCS of the meta-UCB algorithm when applied to a generic R\&S problem. This analysis lays the foundation for further understanding the algorithm’s performance in solving large-scale, non-sub-Gaussian problems. 
% Despite the clean and intuitive structure of the meta-UCB algorithm (Algorithm~\ref{algo: metaUCB}), characterizing its PCS is nontrivial due to its sequential and adaptive nature.
Prior studies on UCB algorithms have attempted similar analyses, but they often fail to yield meaningful insights for large-scale problems—resulting in uninformative bounds (e.g., lower bounds of \(-\infty\) for the PCS as \(k\to\infty\))—or rely heavily on sub-Gaussian assumptions \citep{audibert2010best, jamieson2014lil}. \textcolor{black}{We approach the analysis from a boundary-crossing perspective inspired by \citet{li2023surprising}, where this perspective was used to examine the sample means of different alternatives to understand the performance of greedy algorithms for Gaussian R\&S.}
In Section~\ref{subsec: boundary-crossing}, we introduce the boundary-crossing perspective to review the sequential sampling process through UCB value processes and derive a neat and distribution-free PCS lower bound for the meta-UCB algorithm. This PCS analysis may highlight the role of the exploration bonus in R\&S, which we further explore in Section~\ref{subsec: insight}.
% Jeff comment: You may want to mention what Li et al. (2025) is about to differentiate it from this paper. The current treatment is a bit strange. Inspired by \cite{li2023surprising}

\subsection{A Boundary-Crossing Perspective}
\label{subsec: boundary-crossing}

\begin{figure}[htp]
         \FIGURE
    {\includegraphics[width=1\linewidth]{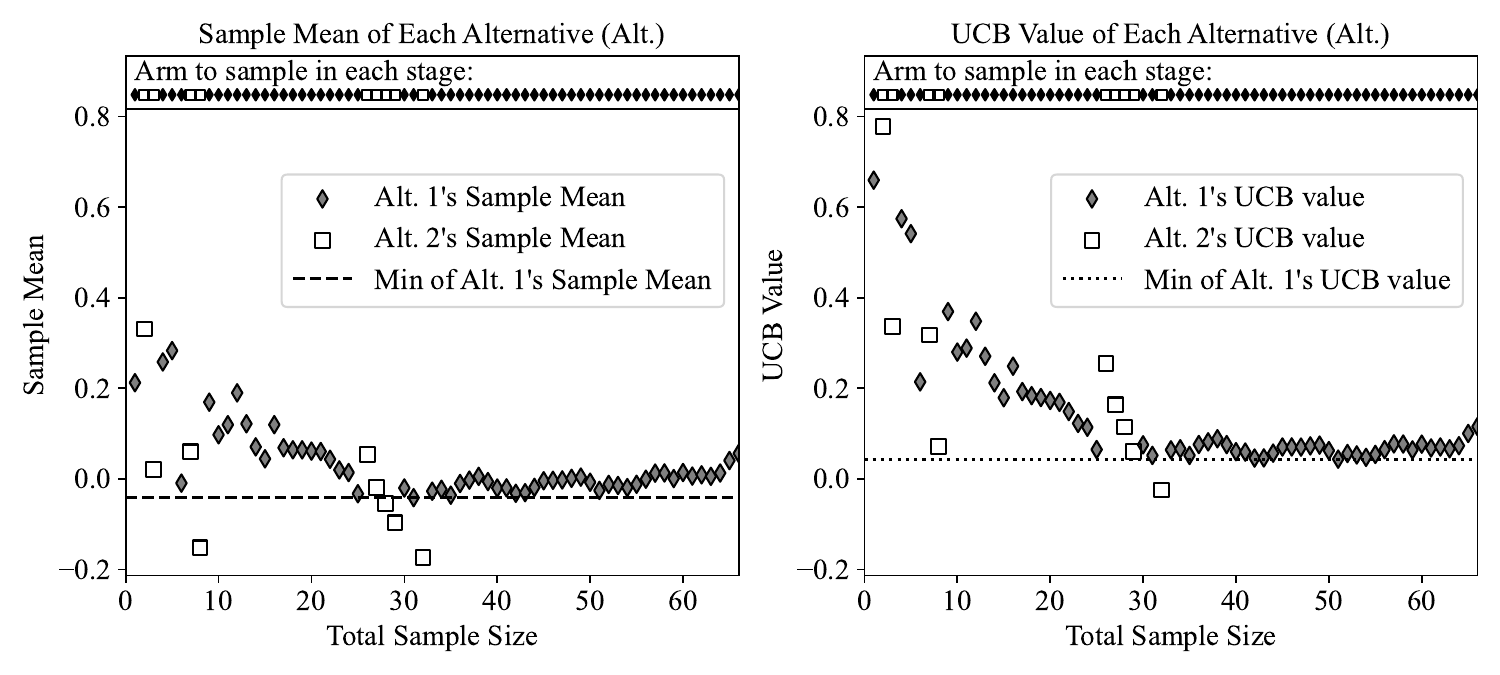}}
     {A Sampling Process of the UCB algorithm with $U_i(n)=\bar X_i(n) + \sqrt{0.2/n}$ for a Problem with 2 Alternatives.\label{fig:UCB_process}}{}
\end{figure}

To characterize the PCS of the meta-UCB algorithm, we adopt a boundary-crossing perspective to analyze its sampling process. Under this perspective, we attempt to regard the best alternative as a \textit{boundary} for all non-best alternatives and model the sampling dynamics of each non-best alternative as a \textit{boundary-crossing process} relative to the boundary. To illustrate the key ideas, let us consider a concrete example:  the UCBE algorithm with  UCB value $U_i(n)=\bar X_i(n) + \sqrt{0.2/n}$, applied to solve an example problem with only two alternatives where $X_1 \sim \text{Normal}(0.1, 0.25)$ and $X_2 \sim \text{Normal}(0, 0.25)$. Figure \ref{fig:UCB_process} shows a realized sampling process of the UCB algorithm, illustrating how the sample means and UCB values for both alternatives evolve as the total sample size increases.

 %where $X_1 \sim \text{Normal}(0.1, 0.25)$ and $X_2 \sim \text{Normal}(0, 0.25)$.

Figure~\ref{fig:UCB_process} reveals a critical observation: \emph{once the UCB value of alternative 2 falls below the minimum of alternative 1's UCB values, it may never be selected and sampled again}. This phenomenon arises from the decoupled structure of UCB values across different alternatives, as highlighted in Assumption~\ref{assu: bonusfunction}. Since the UCB values are computed independently for each alternative, sampling alternative 1 does not affect the UCB value of alternative 2. As a result, regardless of how many additional observations alternative 1 receives, alternative 2 remains uncompetitive once its UCB value falls below the minimum of alternative 1's UCB values. In this sense, the minimum UCB value of the best alternative effectively forms a boundary for all non-best alternatives. This observation motivates a boundary-crossing perspective for analyzing the meta-UCB algorithm. 

To formalize the observation, we define the stochastic process \(\{U_i(n): n = 1, 2, \dots\}\) as alternative \(i\)'s \textit{UCB value process}. These UCB value processes for all alternatives collectively determine the meta-UCB algorithm's sampling process across the alternatives. For the best alternative, we define  
\[
U_1^* = \inf_{n \geq 1} U_1(n)
\]
as the minimum of its UCB value process. Notice that $U_1^*$ is a random variable defined over the entire UCB value process from $n=1$ to $\infty$, and it is independent of the sampling behavior of the algorithm.  
In cases with $U_1(n) > \mu_1$ for all $n \geq 1$ and $U_1(n)\to\mu_1$ (see Section \ref{subsec: no_IZ} for examples), $U_1^*=\mu_1$. 
 % since $\lim_{n\to\infty} U_i(n)=\mu_i$ almost surely
Inspired by Figure \ref{fig:UCB_process}, we propose using \(U_1^*\) as the boundary for all non-best alternatives. From this boundary-crossing perspective, we arrive at the following property of the algorithm.  

\begin{property}
\label{lem: boundary-crossing}
During the sampling process of the meta-UCB algorithm, for every non-best alternative \(i=2,\cdots,k\), \textit{once its UCB value \(U_i(n)\) falls below \(U_1^*\), it will be dominated by alternative 1 and never be sampled again}.
\end{property}

% Figure 1 in the following visualizes this observation on an example pure exploration problem with 3 alternatives. 
% The boundary-crossing perspective established in 
Property \ref{lem: boundary-crossing} is particularly useful, as it allows us to directly obtain an upper bound on the allocated sample size of each non-best alternative. This result is summarized in the following property.

\begin{property} 
\label{lem: bound_size}
For the meta-UCB algorithm, regardless of the total sampling budget $B$, the total number of observations allocated to each non-best alternative $i=2, \cdots, k$, denoted by $n_i(B)$, satisfies
\begin{eqnarray}
    \label{eq: bound_size}
    n_i(B) \leq  T_i^f(U^*_1) := \inf\{n  \geq 1: U_i(n) < U^*_1\}, \quad a.s.,
\end{eqnarray}
where $T_i^f(U^*_1)$ denotes the boundary-crossing time of alternative $i$ regarding the boundary $U^*_1$.
\end{property}

% This is something we mentioned in the top-m paper
% Jeff comment: This is confusing! U_1^* is a random variable, so is n_i(B). You need to mention that they are measured in the same sample path. Also, as U_1^* may be low, it is possible that for some sample paths, the bound may be infinity. 
\begin{remark}
\textcolor{black}{
Both $U_1^*$ and $n_i(B)$ are random variables defined on the same sample path, which is composed of the UCB value processes \(\{U_i(n): n = 1, 2, \dots\}\) of all alternatives. The boundary-crossing time $T_i^f(U_1^*)$ may be infinite when $U_1^*$ is no larger than the infimum of alternative $i$'s UCB values; by contrast, $n_i(B)$ is always finite for every finite budget $B$. This possibility for the boundary-crossing time does not affect the subsequent PCS analysis.} 
% We will focus on scenarios where such cases do not occur in the following sections.
\end{remark}

With this property, we are ready to study the PCS of the meta-UCB algorithm. Recall that the algorithm selects an alternative with the largest sample size when the total sampling budget is exhausted. Regardless of how ties are broken, strict dominance of alternative 1's terminal sample size is sufficient for correct selection. Thus,
\begin{eqnarray}
\label{eq: PCS_form}
    \mathrm{PCS} \geq \Pr \left\{n_1(B) > \max_{i=2, \cdots, k} n_i(B) \right\} = \Pr \left\{B  >  \max_{i=2, \cdots, k} n_i(B) + \sum_{i=2}^k n_i(B)  \right\}.
\end{eqnarray}
Combining Equations \eqref{eq: bound_size} and \eqref{eq: PCS_form}, we conclude that under a total sampling budget $B$, the PCS of the meta-UCB algorithm satisfies
\begin{eqnarray}
    \label{eq: PCS_bound}
    \mathrm{PCS} \geq \Pr \left\{B  > \max_{i=2, \cdots, k} T_i^f(U^*_1) +  \sum_{i=2}^k T_i^f(U^*_1)\right\} \geq \Pr \left\{B  > 2\sum_{i=2}^k T_i^f(U^*_1)\right\}.
\end{eqnarray} 

Equation~\eqref{eq: PCS_bound} presents a lower bound on the PCS for the meta-UCB algorithm. Notably, this result is distribution-free. The bound depends on the distributional information of the problem only through the boundary-crossing times. Moreover, the PCS bound holds for any problem instance with a unique best alternative and applies to all UCB algorithms that satisfy Assumption~\ref{assu: bonusfunction}. This distribution-free PCS lower bound sets the stage for our subsequent analysis of the meta-UCB algorithm—and therefore a class of UCB algorithms—in solving R\&S problems with general performance distributions. These analyses are developed in Section~\ref{sec: sampleoptimality}.

% Do we really need to consider the difference? we don't need to introduce greedy. 
% We would like to emphasize that our boundary-crossing analysis differs significantly from that of \cite{li2023surprising} (LFH).
% First, we focus on the UCB processes to analyze the meta-UCB algorithm, whereas LFH analyze the sample mean process in the context of sequential greedy algorithms. Second, we use the largest sample size as the selection criterion, while LFH use the largest sample mean as the selection standard in their greedy algorithms. These distinctions make our PCS analysis not only simpler than that in LFH but also more suitable for UCB algorithms.

% Highlight why using sample mean is not a good choice. Some materials in EC. \textcolor{black}{highlight distribution-free}

To the best of our knowledge, Equation~\eqref{eq: PCS_bound} provides the first PCS lower bound for UCB algorithms that is expressed in terms of boundary-crossing times. Both the simplicity of the bound and the elegance of its derivation are noteworthy. From  Equations \eqref{eq: PCS_form} and \eqref{eq: PCS_bound} we can see that a key contributor to this simplicity is our choice of selection criteria: let the algorithm select the alternative with the largest sample size at the end of the sampling process. While other selection criteria—such as the largest sample mean or UCB value—could be considered, we leave the analysis of those alternatives to future work. Hence, we focus on the largest sample size criterion in this paper.

\subsection{Effect of Exploration Bonus}
\label{subsec: insight}
The PCS analysis above provides an alternative lens through which to understand the effect of the exploration bonus function in UCB algorithms. To simplify the presentation, in the following we focus on the special case with $k=2$. The insights gained in this two-alternative scenario can be easily generalized to cases with $k>2$. Directly from Property \ref{lem: bound_size}, we have
\begin{eqnarray}
    \label{eq: PCS_bound_2}
    \Pr \left\{ n_1(B) > n_2(B)  \text{ when } B > 2 T_2^f(U^*_1)\right\} = 1,
\end{eqnarray}
which means that $2 T_2^f(U^*_1)$ can serve as an upper bound for the required sampling budget to ensure a correct selection.
To further expose the effect of $f$ on $T_2^f(U^*_1)$, we introduce a constant $\gamma \in [0, \infty)$ to define a constant boundary $\mu_1 - \gamma$. We then consider the associated boundary-crossing time  
\begin{eqnarray*}
    T_2^f(\mu_1 - \gamma) := \inf \{n  \geq 1: U_2(n) < \mu_1 - \gamma \} = \inf \{n  \geq 1: \bar{X}_2(n) + f(n)< \mu_1 - \gamma \}.
\end{eqnarray*}
Conditional on the event $\left\{ U^*_1 > \mu_1 - \gamma\right\}$, we know that $T_2^f(\mu_1 - \gamma) \geq T_2^f(U^*_1)$. Consequently,
\begin{eqnarray}
\label{eq: PCS_bound_3}
    \notag & & \Pr \left\{n_1(B) > n_2(B)  \text{ when } B > 2 T_2^f(\mu_1 - \gamma)  \right\} \\
     \notag & & \quad \geq  \Pr \left\{n_1(B) > n_2(B)  \text{ when } B > 2 T_2^f(\mu_1 - \gamma)  \mid U^*_1 \geq   \mu_1 - \gamma \right\} \Pr \left\{U^*_1 \geq   \mu_1 - \gamma\right\} \\ 
     \notag & & \quad \geq \Pr \left\{n_1(B) > n_2(B)  \text{ when } B > 2 T_2^f(U_1^*)  \mid U^*_1 \geq   \mu_1 - \gamma \right\} \Pr \left\{U^*_1 \geq   \mu_1 - \gamma\right\} \\ 
     & & \quad  = \Pr \left\{U^*_1 \geq   \mu_1 - \gamma\right\} = \Pr \left\{\bar{X}_1(n)  + f(n) \geq \mu_1 - \gamma , \,  \forall n \geq 1\right\} =: P_f,
\end{eqnarray}
where the first equality arises from Equation \eqref{eq: PCS_bound_2}.
Equation \eqref{eq: PCS_bound_3} is noteworthy as it suggests that \( 2 T_2^f(\mu_1 - \gamma) \) can be viewed as an upper bound for the meta-UCB algorithm's required sampling budget to guarantee a PCS of at least \( P_f \).  This result highlights the dual role played by the non-negative exploration bonus function $f$ in UCB algorithms. On the one hand, it controls exploration. Increasing the exploration bonus amplifies the optimism in the UCB values, thereby improving the chance of correctly identifying the best alternative $P_f$. This aligns with the conventional understanding that the exploration bonus helps promote better selection. On the other hand, a higher exploration bonus also increases the boundary-crossing time $T_2^f(\mu_1 - \gamma)$, which implies a larger required sampling budget. In this light, boundary-crossing times offer a tangible way to quantify the cost of using the exploration bonus.

% \begin{figure}[htbp]
%      \centering
%      \begin{subfigure}[b]{0.33\textwidth}
%          \centering
%          \includegraphics[width=\textwidth]{compare1.png}
%          \caption{}
%          \label{fig:compare1}
%      \end{subfigure}
%      \hfill
%      \begin{subfigure}[b]{0.32\textwidth}
%          \centering
%          \includegraphics[width=\textwidth]{compare2.png}
%          \caption{}
%          \label{fig:compare2}
%      \end{subfigure}
%      \hfill
%      \begin{subfigure}[b]{0.33\textwidth}
%          \centering
%          \includegraphics[width=\textwidth]{compare3.png}
%          \caption{}
%          \label{fig:compare3}
%      \end{subfigure}
%         \caption{A comparison between the bonus functions $f_1(n) = n^{-1/2}$ and $f_2(n) = n^{-1/3}$  on an example problem with $k=2$}
%         \label{fig:three graphs}
% \end{figure}

% \begin{figure}[htbp]
%     \center{
%     \includegraphics[width=0.28\textwidth]{compare1.png}
%     \includegraphics[width=0.26 \textwidth]{compare2.png}
%     \includegraphics[width=0.28\textwidth]{compare3.png}}
%     \caption{Comparing $f_1(n) = n^{-1/2}$ and $f_2(n) = n^{-1/3}$, their corespondent boundaries  and the first boundary-crossing times}
%     \label{fig:Example2}
% \end{figure}

\section{Sample Optimality of the Meta-UCB Algorithm}
\label{sec: sampleoptimality}

By leveraging the distribution-free PCS lower bound in Equation~\eqref{eq: PCS_bound}, we now investigate the properties of the meta-UCB algorithm in solving non-sub-Gaussian R\&S problems. In Section~\ref{subsec: distributionfreeanalysis}, we formalize sufficient conditions for sample optimality. In Section~\ref{subsec: assumptions}, we state a uniform variance condition and the IZ condition. Section~\ref{subsec: second_moment_iz} then establishes IZ sample optimality under these conditions.

\subsection{Sufficient Conditions for Sample Optimality}
\label{subsec: distributionfreeanalysis}
Recall from Equation \eqref{eq: PCS_bound} that the PCS of the meta-UCB algorithm satisfies
\begin{eqnarray}
\label{eq: PCS_bound_22}
    \mathrm{PCS} \geq \Pr \left\{B  > 2\sum_{i=2}^k T_i^f(U^*_1)\right\}.
\end{eqnarray}
To study the sample optimality, we need to drive $k\to\infty$ and analyze the behavior of the sum of $k-1$ boundary-crossing times. A key technical challenge arises from the dependence across these times due to the shared random boundary $U^*_1$. The conditioning arguments in Section \ref{subsec: insight} provide a way to get around this. Consider the event $\{U^*_1 \geq \mu_1 - \gamma_0 \}$ for some $\gamma_0  \in [0, \infty)$. Then Equation~\eqref{eq: PCS_bound_22} implies
\begin{eqnarray}
    \label{eq: PCS_bound_event}
    \notag \mathrm{PCS} & \geq & \Pr \left\{B >   2 \sum_{i=2}^k T_i^f(U^*_1) \Big| U^*_1 \geq \mu_1 - \gamma_0 \right\} \Pr \left\{U^*_1 \geq \mu_1 - \gamma_0  \right\} \\ 
    & \geq & \Pr \left\{B > 2\sum_{i=2}^k T_i^f(\mu_1 - \gamma_0 )\right\} \Pr \left\{U^*_1 \geq \mu_1 - \gamma_0  \right\}.
\end{eqnarray}
This decomposition effectively decouples the best alternative from the non-best ones, and the boundary-crossing times $T_i^f(\mu_1 - \gamma_0 )$ are mutually independent as now the boundary is a constant. From Equation \eqref{eq: PCS_bound_event}, considering the asymptotic regime of letting $k\to\infty$, we can see that the following conditions would be sufficient to show the sample optimality. The alternatives' means and distributions may depend on $k$; we suppress this dependence to avoid cumbersome notation.
\begin{condition}
\label{argu: 1}
% = \Pr \left\{\bar X_1(n) + f(n) \geq \mu_1 - \gamma_0, \, \forall n \geq 1  \right\}
For some fixed $\gamma_0\in[0,\infty)$,
\[
\liminf_{k\to\infty}\Pr\left\{U_1^*\geq\mu_1-\gamma_0\right\}
=\liminf_{k\to\infty}\Pr\left\{U_1(n)\geq\mu_1-\gamma_0,\ \forall n\geq1\right\}>0.
\]
\end{condition}
\begin{condition}
\label{argu: 2}
For the $\gamma_0  \in [0, \infty)$ satisfying Condition \ref{argu: 1}, the sum $\sum_{i=2}^k T_i^f(\mu_1 - \gamma_0 )$ grows at most linearly in $k$ in probability, i.e., there exists a constant $c(\gamma_0)>0$ such that, for every $\varepsilon>0$,
\[
\lim_{k\to\infty}
\Pr\left\{\sum_{i=2}^k T_i^f(\mu_1 - \gamma_0 )
\leq \bigl(c(\gamma_0)+\varepsilon\bigr)k\right\}=1.
\]
\end{condition}

Under these two conditions, choose any $c_B>2c(\gamma_0)$ and let $B=c_Bk$. Taking $\varepsilon\in(0,c_B/2-c(\gamma_0))$ in Condition~\ref{argu: 2} and using Equation~\eqref{eq: PCS_bound_event} gives
\[
\liminf_{k\to\infty} \mathrm{PCS}
\geq \liminf_{k\to\infty}\Pr \left\{U^*_1 \geq \mu_1 - \gamma_0  \right\}>0,
\]
which satisfies the notion of sample optimality as defined in Definition~\ref{def: rate_optimality_PCS}. Therefore, if these two conditions can be verified, the sample optimality of the meta-UCB algorithm would follow.
 % and highlights their intended behavior: maintaining optimism about the true best alternative
Condition \ref{argu: 1} requires that the UCB value of the best alternative (alternative 1) remains above $\mu_1 - \gamma_0$ with a probability bounded away from zero uniformly for all sufficiently large $k$. This aligns with the interpretation of upper confidence bounds in UCB algorithms (particularly in the case with $\gamma_0 = 0$). Notably, this condition only concerns alternative 1 and does not impose requirements on the other alternatives. This is intuitive—ensuring a uniformly non-vanishing confidence level across all $k$ alternatives becomes increasingly infeasible as $k \to \infty$, due to multiplicity effects. Condition \ref{argu: 2} is a concentration requirement for the aggregate boundary-crossing cost. Its in-probability formulation accommodates triangular arrays in which the alternatives' distributions may change with $k$. Both conditions ultimately depend on the distributional properties of the alternatives. 

We will repeatedly use the following elementary consequence of Chebyshev's inequality to verify the aggregate concentration condition. The proof is provided in Section~\ref{subsec: proof_lem_linear_mean_subquadratic_variance}.

\begin{lemma}[Linear-Mean, Subquadratic-Variance Concentration]
\label{lem: linear_mean_subquadratic_variance}
Let $(S_k)_{k\geq1}$ be any sequence of random variables. Suppose that, for some constant $C<\infty$, $\E[S_k]\leq Ck$ for all sufficiently large $k$ and $\operatorname{Var}(S_k)=o(k^2)$. 
Then, for every $\varepsilon>0$,
$
\lim_{k\to\infty}
\Pr\left\{S_k\geq(C+\varepsilon)k\right\}=0.
$
\end{lemma}

% Lemma~\ref{lem: linear_mean_subquadratic_variance} concerns an arbitrary sequence $(S_k)$; it is not restricted to triangular arrays. In a row-wise application, independence within a row is used only to verify the variance condition for the row sum. The rows themselves need not be nested or identically distributed.

\subsection{Variance and Indifference-Zone Conditions}
\label{subsec: assumptions}

For the remainder of this section, we impose the following uniform variance bound and IZ gap condition.

\begin{assumption}[Uniformly Bounded Variances]
    \label{assu: bounded_variance}
For some $\bar\sigma^2<\infty$, $\operatorname{Var}(X_i)\leq\bar\sigma^2$ for every $k$ and every $i=1,\ldots,k$. Write $\sigma_i^2:=\operatorname{Var}(X_i)$.
\end{assumption}

Assumption~\ref{assu: bounded_variance} is mild in the sense that variances may differ across alternatives; it only rules out variances that grow without bound as $k$ increases.  We impose no additional restrictions on the distributional families. We allow the alternatives to follow different distribution families.

\begin{assumption}[Indifference-Zone Formulation]
\label{assu: iz}
For some $\gamma>0$, $\mu_1-\max_{2\leq i\leq k}\mu_i\geq\gamma$ for every $k$.
\end{assumption}

This is the conventional indifference-zone (IZ) formulation in the ranking and selection literature \citep{bechhofer1954single, kim2006selecting}; it is also known as the sparse configuration in the best arm identification literature \citep{jamieson2013finding}. The constant mean gap prevents some alternatives from becoming nearly indistinguishable from the best. Recent studies of large-scale ranking and selection use this formulation to study sample optimality for the PCS (e.g., \citealt{zhong2022knockout}). We follow these works in this section and discuss scenarios where the IZ formulation fails later in Section \ref{subsec: no_IZ}. Neither $\gamma$ nor the variance bound $\bar\sigma^2$ is an input of the meta-UCB algorithm; they determine only a sufficient linear-budget constant.

\subsection{Sample Optimality}
\label{subsec: second_moment_iz}
To demonstrate the sample optimality of the meta-UCB algorithm, we now examine Conditions~\ref{argu: 1} and~\ref{argu: 2}. First, Assumptions~\ref{assu: bounded_variance} and~\ref{assu: iz} imply the following lemma. It shows that the IZ buffer makes Condition~\ref{argu: 1} hold for every $\gamma_0\in(0,\gamma)$, regardless of the distributional form of the best alternative. The proof is included in Section \ref{subsec: proof_lem_pcs_bound}.

\begin{lemma}
\label{lem: PCS_bound}
Suppose that Assumptions~\ref{assu: bounded_variance} and~\ref{assu: iz} hold. Then, for any bonus function $f$ satisfying Assumption \ref{assu: bonusfunction} and for any $\gamma_0 \in (0, \gamma)$, we have 
\begin{eqnarray*}
% = \Pr \left\{\forall n \geq 1, \bar X_1(n) \geq \mu_1 - \gamma_0_0 - f(n) \right\}
 % \geq \Pr \left\{\forall n \geq 1, \bar X_1(n) \geq \mu_1 - \gamma_0 \right\}
       \Pr \left\{U^*_1 \geq \mu_1 - \gamma_0  \right\}   \geq \exp \left(-\frac{\pi^2 \sigma_1^2}{6\gamma_0 ^2}\right) > 0.
\end{eqnarray*}
\end{lemma}

In the remainder of this section, we focus on verifying Condition~\ref{argu: 2}, which concerns the properties of the boundary-crossing times of the non-best alternatives $T_i^f(\mu_1 - \gamma_0)$. These times can be rewritten as
\begin{eqnarray}
\label{eq: bc_transform}
T_i^f(\mu_1 - \gamma_0) 
& = & \inf\left\{n  \geq 1: \bar X_i(n) + f(n) < \mu_1-\gamma_0\right\} = \inf\left\{n  \geq 1: \bar X_i(n) < \mu_1 - \gamma_0 - f(n)\right\},
\end{eqnarray}
which denotes the first boundary-crossing time of the sample mean $\bar X_i(n)$ for the boundary $\mu_1-\gamma_0-f(n)$. Analyzing this boundary-crossing time is challenging, since the boundary depends on the general (most likely non-linear) function $f$ that satisfies Assumption \ref{assu: bonusfunction}.
% , which facilitates the analysis by transforming the boundary-crossing condition.

\clearpage
To address this, we introduce the following lemma.

\begin{lemma}
\label{lem: move}
    Let $X_1, X_2, \cdots$ be a sequence of i.i.d. random variables with zero mean and $\bar{X}(n) = \frac{1}{n} \sum_{j=1}^n X_j$. For any bonus function $f$ satisfying Assumption \ref{assu: bonusfunction} and any constant $b>0$, let $n^f(b)$ denote a positive integer such that $f(n) \leq b$ for all $n\geq n^f(b)$.
    %be the \textit{last exit time} for the region $[0, b]$ and let $n^{f}(b) = n(b)+1$ so that for all $n \geq n^{f}(b)$, $f(n) \leq b$. 
     Then
     \begin{eqnarray*}
                          \inf\left\{n  \geq 1: \bar X(n) < b - f(n)\right\} \leq  \inf \left\{ n \geq  n^{f}(b/2): \bar X(n) < b/2\right\} := \bar T^f(b/2) \quad  a.s. 
                    \end{eqnarray*}
\end{lemma}
The proof is provided in Section~\ref{subsec: proof_lem_move}.

Lemma~\ref{lem: move} is particularly useful. It shifts the exploration bonus function $f$ from the boundary to the starting point of the boundary-crossing process. As we will show below, this transformation facilitates a more tractable analysis.

% See Figure \ref{fig:bc_time} for a visualized example.

%                     \begin{figure}[htbp]
%     \centering
%     \includegraphics[width=0.4 \textwidth]{illustration.pdf}
%     \caption{Transformation between boundary-crossing times}
%     \label{fig:bc_time}
% \end{figure}

The following lemma provides first- and second-moment bounds for a delayed boundary-crossing time. Its proof is provided in Section~\ref{subsec: proof_lem_finite_variance_crossing}.

\begin{lemma}[Finite-Variance Crossing Time]
\label{lem: finite_variance_crossing}
Let $Y_1,Y_2,\ldots$ be i.i.d. random variables with $\E[Y_1]=0$ and $\operatorname{Var}(Y_1)=\sigma^2<\infty$. For $b>0$ and a positive integer $n_0$, define
$
H(b;n_0)=\inf\left\{n\geq n_0:\frac1n\sum_{j=1}^nY_j<b\right\}.
$
Then
\begin{eqnarray}
\label{eq: finite_variance_crossing_bounds}
\E[H(b;n_0)]
&\leq&n_0\exp\left(\frac{\pi^2\sigma^2}{6n_0b^2}\right),\nonumber\\
\E[H(b;n_0)^2]
&\leq&n_0^2\left(2+\frac{4\sigma^2}{n_0b^2}\right)
\exp\left(\frac{\pi^2\sigma^2}{6n_0b^2}\right).
\end{eqnarray}
\end{lemma}

Fix $\gamma_0\in(0,\gamma)$. For every non-best alternative $i$, Assumption~\ref{assu: iz} gives $\mu_1-\gamma_0-\mu_i=(\mu_1-\mu_i)-\gamma_0\geq\gamma-\gamma_0>0$. Therefore, boundary monotonicity and Lemma~\ref{lem: move} give
\begin{eqnarray*}
T_i^f(\mu_1-\gamma_0)
&=&\inf\left\{n\geq1:\bar X_i(n)-\mu_i
<\mu_1-\gamma_0-\mu_i-f(n)\right\}\\
&\leq&\inf\left\{n\geq1:\bar X_i(n)-\mu_i
<\gamma-\gamma_0-f(n)\right\}\\
&\leq&\inf\left\{n\geq n^f((\gamma-\gamma_0)/2):
\bar X_i(n)-\mu_i<\frac{\gamma-\gamma_0}{2}\right\}.
\end{eqnarray*}
% For the first inequality, replacing $\mu_1-\gamma_0-\mu_i$ by the smaller boundary $\gamma-\gamma_0$ can only delay the first crossing and therefore gives an upper bound. For the second, whenever $n\geq n^f((\gamma-\gamma_0)/2)$, the definition of $n^f$ gives $f(n)\leq(\gamma-\gamma_0)/2$, so $\gamma-\gamma_0-f(n)\geq(\gamma-\gamma_0)/2$. Thus, the first time after $n^f((\gamma-\gamma_0)/2)$ that the centered sample mean falls below the half-gap is a valid crossing time for the preceding boundary. This is precisely Lemma~\ref{lem: move}.

Set $n_0=n^f((\gamma-\gamma_0)/2)$. Under Assumption~\ref{assu: bounded_variance}, Lemma~\ref{lem: finite_variance_crossing}, with crossing level $(\gamma-\gamma_0)/2$ and starting time $n_0$, now yields the uniform bounds
\begin{eqnarray}
\label{eq: iz_crossing_constants}
\E\left[T_i^f(\mu_1-\gamma_0)\right]&\leq&C_2^f(\gamma_0)
:=n_0\exp\left(\frac{2\pi^2\bar\sigma^2}{3n_0(\gamma-\gamma_0)^2}\right),\nonumber\\
\E\left[\left(T_i^f(\mu_1-\gamma_0)\right)^2\right]&\leq&D_2^f(\gamma_0)
:=n_0^2\left(2+\frac{16\bar\sigma^2}{n_0(\gamma-\gamma_0)^2}\right)
\exp\left(\frac{2\pi^2\bar\sigma^2}{3n_0(\gamma-\gamma_0)^2}\right).
\end{eqnarray}

Together with independence across alternatives, these bounds verify the hypotheses of Lemma~\ref{lem: linear_mean_subquadratic_variance} for the aggregate crossing cost, and hence Condition~\ref{argu: 2}. Combining this with Lemma~\ref{lem: PCS_bound} yields the following theorem; its proof is provided in Section~\ref{subsec: proof_thm_iz_second_moment}.

\begin{theorem}[IZ Sample Optimality under Uniformly Bounded Variances]
\label{thm: iz_second_moment}
Suppose that Assumptions~\ref{assu: bonusfunction}, \ref{assu: bounded_variance}, and~\ref{assu: iz} hold. Fix any $\gamma_0\in(0,\gamma)$ and define $C_2^f(\gamma_0)$ as in Equation~\eqref{eq: iz_crossing_constants}. If $B=ck$ for some $c>2C_2^f(\gamma_0)$, then
\[
\liminf_{k\to\infty}\mathrm{PCS}
\geq
\exp\left(-\frac{\pi^2\bar\sigma^2}{6\gamma_0^2}\right)>0.
\]
Consequently, the meta-UCB algorithm is sample optimal.
\end{theorem}

Theorem~\ref{thm: iz_second_moment} confirms the sample optimality of the meta-UCB algorithm, thereby establishing the sample optimality of all UCB algorithms that satisfy Assumption~\ref{assu: bonusfunction} for large-scale R\&S. This result demonstrates that UCB algorithms can be sample optimal without relying on Gaussian or sub-Gaussian assumptions. Besides establishing sample optimality, Theorem ~\ref{thm: iz_second_moment}provides a lower bound for the PCS in the limit $k\to\infty$. This lower bound is defined on the distributional property of the best alternative and a constant $\gamma_0$ purely determined by the means and variances of all alternatives, sampling budget, and the exploration bonus function.

\section{Extensions and Discussions}
\label{sec: extensions}

Theorem~\ref{thm: iz_second_moment} is established under the indifference-zone formulation in Assumption~\ref{assu: iz}, which requires a constant mean gap between the best alternative and all others. In this section, we explore extensions beyond this scope. Section~\ref{subsec: no_IZ} considers polynomially shrinking gaps under Assumption~\ref{assu: bounded_variance}. Section~\ref{subsec: more_general} discusses broader classes of UCB values beyond Assumption~\ref{assu: bonusfunction}.
 % and the mean gaps may shrink with $k$

\subsection{Moving Beyond the Indifference-Zone Formulation}
\label{subsec: no_IZ}

The  IZ formulation in Assumption~\ref{assu: iz} may be argued to be impractical for large-scale problems, where the mean differences between the best alternative and some non-best ones may shrink as $k$ grows \citep{ni2017efficient}. Motivated by this, we examine whether the meta-UCB algorithm can achieve the sample optimality in scenarios where Assumption~\ref{assu: iz} does not hold. To model such scenarios, we consider a representative configuration in which the mean gap between the best alternative and alternative $i$, denoted $\Delta_i = \mu_1 - \mu_i$, shrinks at a polynomial rate, namely at the rate of $\mathcal{O}(k^{-\beta})$ for some $\beta > 0$. Inspired by \citet{jamieson2013finding}, we adopt the following mean configuration:
\begin{equation}
\label{eq: non-iz-config}
\mu_i = \mu_1 - (i/k)^{\beta}  \quad \text{for } i = 2, \dots, k.
\end{equation}
This formulation captures two important features commonly observed in large-scale problems where Assumption~\ref{assu: iz} fails to hold:
(1) the means are spread across a range and become increasingly dense as the number of alternatives grows;
(2) the smallest mean gap $\Delta_2$ diminishes as $k \to \infty$. The parameter $\beta$ intuitively characterizes the hardness of the selection problem: a larger $\beta$ implies that the non-best means are more tightly clustered near the best mean $\mu_1$, making it more difficult to identify the true best alternative. When $\beta = 0$, the configuration satisfies Assumption~\ref{assu: iz}, recovering the ``easiest'' IZ formulation.
%at the rate $\mathcal{O}(k^{-\beta})$
      % This configuration provides a benchmark for comparing different algorithms' performance. 

Under this configuration, for the meta-UCB algorithm to achieve the sample optimality, the following two requirements—similar to those discussed in Section~\ref{subsec: distributionfreeanalysis}—serve as sufficient conditions.  Intuitively, these two conditions can be regarded as adjustments to Conditions \ref{argu: 1} and \ref{argu: 2}, achieved by enforcing $\gamma = 0$.
\begin{condition}
\label{argu: 3}
$\liminf_{k\to\infty}\Pr \left\{U^*_1 \geq \mu_1  \right\} > 0$.
\end{condition}
\begin{condition}
\label{argu: 4}
There exists a constant $c>0$ such that, for every $\varepsilon>0$,
\[
\lim_{k\to\infty}
\Pr\left\{\sum_{i=2}^kT_i^f(\mu_1)\leq(c+\varepsilon)k\right\}=1.
\]
\end{condition}

\subsubsection{Satisfaction of Condition~\ref{argu: 3}.}
To ensure Condition~\ref{argu: 3}, it is enough to choose the bonus so that, for some fixed $\alpha\in(0,1)$,
\begin{eqnarray}
\label{eq: confidence}
\Pr\{U_1^*\geq\mu_1\}
=\Pr\left\{\bar X_1(n)+f(n)\geq\mu_1\text{ for every }n\geq1\right\}
\geq1-\alpha.
\end{eqnarray}
Thus the UCB process of the best alternative must form a one-sided time-uniform confidence sequence. This is not automatic under Assumption~\ref{assu: bonusfunction}; in particular, a bonus may vanish too quickly to protect the sample mean over an infinite time horizon. Under the uniform variance bound, the following construction provides such a confidence sequence. Its proof is provided in Section~\ref{subsec: proof_lem_variance_CS}.

\begin{lemma}[Anytime Bonus under Bounded Variance]
\label{lem: variance_CS}
Suppose that $\operatorname{Var}(X_1)\leq\bar\sigma^2$. Fix $\alpha\in(0,1)$ and choose $A\geq\pi\bar\sigma/\sqrt{3\alpha}$.
Then the exploration bonus
\begin{equation}
\label{eq: log_sqrt_bonus}
f_A(n)=A\frac{\log_2(2n)}{\sqrt n},\qquad n\geq1,
\end{equation}
satisfies Assumption~\ref{assu: bonusfunction} and Equation~\eqref{eq: confidence}.
\end{lemma}
 
This construction is near optimal for the sample size $n$, as the law of iterated logarithm shows that the sample mean of any random variable with finite variance scales at most as $ \mathcal{O}(\sqrt{\log\log n/n})$ asymptotically. 

\subsubsection{Satisfaction of Condition~\ref{argu: 4}.}
We now control the aggregate boundary-crossing cost under the same variance assumption.
\begin{lemma}[Inverse Order of the Logarithmic Square-Root Bonus]
\label{lem: log_bonus_inverse}
For every $A>0$, there exists a positive-integer-valued function $N_A$ on $(0,\infty)$ such that, for every $b>0$, $f_A(n)\leq b$ for all $n\geq N_A(b)$ and
\[
\inf_{b>0} b^2N_A(b)>0.
\]
Moreover, there is a finite constant $\kappa_A$ such that
\[
N_A(b)\leq \kappa_A b^{-2}\left[1+\log(1/b)\right]^2,
\qquad 0<b\leq1/2.
\]
\end{lemma}
The proof is provided in Section~\ref{subsec: proof_lem_log_bonus_inverse}.

Fix a function $N_A$ satisfying Lemma~\ref{lem: log_bonus_inverse}. For the boundary-crossing time $T_i^{f_A}(\mu_1)$ of non-best alternative $i$, put $\Delta_i=\mu_1-\mu_i=(i/k)^\beta$ and $b_i=\Delta_i/2$, and use Lemma~\ref{lem: move} to obtain
\begin{eqnarray}
\label{eq: bc_f}
T_i^{f_A}(\mu_1)
&=&\inf\left\{n\geq1:\bar X_i(n)+f_A(n)<\mu_1\right\}\nonumber\\
&\leq&\inf\left\{n\geq N_A(b_i):
\bar X_i(n)-\mu_i<b_i\right\}
=:\widehat T_{i,k}.
\end{eqnarray}
The next lemma combines Lemmas~\ref{lem: log_bonus_inverse} and~\ref{lem: finite_variance_crossing}. Its proof is provided in Section~\ref{subsec: proof_lem_noniz_second_moment_cost}.

\begin{lemma}[Aggregate Non-IZ Crossing Cost]
\label{lem: noniz_second_moment_cost}
Suppose that Assumption~\ref{assu: bounded_variance} holds, use the bonus in Equation~\eqref{eq: log_sqrt_bonus}, and consider the configuration in Equation~\eqref{eq: non-iz-config} with $0<\beta<1/2$. For the variables $\widehat T_{i,k}$ in Equation~\eqref{eq: bc_f}, there exists a finite constant $D_{\beta,A}$, independent of $k$, such that
\[
\sum_{i=2}^k\E[\widehat T_{i,k}]
\leq D_{\beta,A}k,
\qquad
\sum_{i=2}^k\E[\widehat T_{i,k}^2]
=o(k^2).
\]
\end{lemma}

Lemma~\ref{lem: variance_CS} verifies Condition~\ref{argu: 3}, while Lemmas~\ref{lem: noniz_second_moment_cost} and~\ref{lem: linear_mean_subquadratic_variance}, together with Equation~\eqref{eq: bc_f}, verify Condition~\ref{argu: 4}. Together they yield the following proposition.
 
\begin{proposition}[Non-IZ Sample Optimality under Uniformly Bounded Variances]
\label{prop: non-iz-optimality}
Suppose that Assumption~\ref{assu: bounded_variance} holds and consider the configuration in Equation~\eqref{eq: non-iz-config} with $0<\beta<1/2$. Fix $\alpha\in(0,1)$, choose $A\geq\pi\bar\sigma/\sqrt{3\alpha}$, and run the meta-UCB algorithm with the bonus in Equation~\eqref{eq: log_sqrt_bonus}. Let $D_{\beta,A}$ be the constant in Lemma~\ref{lem: noniz_second_moment_cost}. If $B=ck$ for some $c>2D_{\beta,A}$, then
\[
\liminf_{k\to\infty}\mathrm{PCS}\geq1-\alpha>0.
\]
Consequently, the meta-UCB algorithm is sample optimal.
\end{proposition}

The complete proof is provided in Section~\ref{subsec: proof_prop_noniz_second_moment}. 
Proposition~\ref{prop: non-iz-optimality} is a possibility result—it showcases that the meta-UCB algorithm \emph{can} achieve the sample optimality even without the IZ formulation, for large-scale problems with generally distributed alternatives.
The condition $\beta<1/2$ is exactly what makes the aggregate leading cost
\[
\sum_{i=2}^k\Delta_i^{-2}
=\sum_{i=2}^k(k/i)^{2\beta}
\]
linear in $k$. The logarithmic factor in the bonus does not reduce this range: its inverse contributes only powers of $1+\log(k/i)$, which remain integrable against $x^{-2\beta}$ on $(0,1]$.

\subsection{On the Other UCB Algorithms}
\label{subsec: more_general}

So far, we have focused on UCB algorithms whose UCB values take the form of a sample mean plus a deterministic bonus function of the sample size (i.e., those satisfying Assumption~\ref{assu: bonusfunction}). However, it is natural to ask whether these results may extend to an even broader class of algorithms to allow greater flexibility in how the UCB values can be constructed. For heavy-tailed observations under weak moment conditions, robust mean estimators such as the truncated sample mean or the weighted sample mean and their associated confidence bounds or sequences have proven particularly useful \citep{bubeck2013bandits, glynn2018selecting, wang2023catoni}. In addition, auxiliary sample statistics such as the sample variance can also be incorporated into UCB value constructions \citep{audibert2009exploration}.  
% bubeck2013bandits,

Although our current results do not cover these variants, we conjecture that sample optimality may still hold for a broader class of UCB algorithms, provided that their UCB value construction satisfies a form of \emph{decoupling}—that is, the UCB value for each alternative depends only on its own sample size and sample observations. Under the decoupling condition, the boundary-crossing perspective developed in Section~\ref{subsec: boundary-crossing} remains applicable. Specifically, one can still define a UCB value process $\{U_i(n) : n = 1, 2, \dots\}$ for each alternative $i$ that is independent of that of all other alternatives. This enables the definition of a distribution-free boundary $U_1^*$ for the best alternative, along with corresponding boundary-crossing times $T_i^f(U_1^*)$ for the others.  While the sample optimality analysis would then follow a similar structure as in Section~\ref{sec: sampleoptimality}, verifying the conditions for specific UCB algorithms and distributional assumptions will likely require case-by-case analysis, which we leave for future work. 
% Nonetheless, our exploratory numerical experiments in Section~\ref{subsec: num_more_general} provide empirical support for this conjecture.

%, relying only on two key conditions: (1) the existence of a valid confidence sequence for the best alternative, and (2) the applicability of Kolmogorov’s SLLN for bounding the sum of boundary-crossing times

We also note that there exists another major class of UCB algorithms that do not satisfy the decoupling condition. For example, many algorithms construct UCB values using global statistics, such as the total sample size across all alternatives. A prominent example is the classical UCB1 algorithm of \citet{auer2002finite}. In such cases, the boundary-crossing framework may break down. Interestingly, our experiments in Section~\ref{sec: numerical} show that UCB1 behaves quite differently and may fail to achieve sample optimality. This contrast highlights a fundamental distinction between the two categories of UCB algorithms when used for R\&S.
%, as the UCB value process of each arm is no longer independent of the sampling dynamics of others

\section{Numerical Experiments}
\label{sec: numerical}

In this section, we conduct numerical experiments to examine the theoretical insights and the allocation behavior of representative UCB algorithms. Specifically, in Section~\ref{subsec: num_confs}, we introduce the problem configurations, experimental settings, and algorithms under evaluation. In Section~\ref{subsec: sample_optimal_IZ}, we illustrate behavior consistent with sample optimality under the indifference-zone (IZ) formulation. Section~\ref{subsec: budget_allocation} compares the budget-allocation dynamics of different UCB algorithms.

\subsection{Problem Configurations, Experimental Settings, and Tested UCB Algorithms}
\label{subsec: num_confs}
\subsubsection{Problem Configurations} An R\&S problem configuration specifies the distributional characteristics of all alternatives and serves as a tool for probing the behavior of a given algorithm. To validate our theoretical results, we consider configurations that allow us to examine two key aspects:
(1) how different mean configurations affect algorithm performance, and
(2) how the algorithm behaves under different distributional families.
To meet these goals, we adopt a mean-shifting model to generate problem configurations. We begin with a basic mean-shifting model, in which we first fix the base distribution of a reference random variable $X_0$, and then generate each alternative's performance distribution by shifting the mean as follows:
\begin{equation}
    \label{eq: mean_shifting}
X_i \stackrel{d}{=} 
\begin{cases}
X_0, & \text{if } i = 1, \\\\
X_0 - \gamma - \lambda (i/k)^{\beta}, & \text{if } i = 2, \dots, k.
\end{cases}
\end{equation}
where $\stackrel{d}{=}$ denotes equality in distribution. Here, $\gamma$ is the IZ parameter, $\lambda$ controls the overall spacing of the non-best means, and $\beta$ adjusts how quickly the inferior means decay as $k$ increases. Notice that $X_1, \dots, X_k$ are assumed to be mutually independent. This flexible construction allows us to simulate various scenarios. Parameter values are scenario-specific and described in the relevant subsections. In our experiments, we primarily consider the following two widely studied mean configurations:
\begin{itemize}
\item the Slippage Configuration (SC): Corresponds to $\gamma > 0$ and $\lambda = 0$. This configuration represents a worst-case scenario among all mean structures with the same IZ parameter $\gamma > 0$.
\item the Monotonic Means Configuration (MM): Corresponds to $\lambda > 0$ and $\beta > 0$. This setting is commonly used to test algorithms in more realistic scenarios than SC.
\end{itemize}

Under both the SC and MM configurations, we use finite-variance distributions covered by Theorem~\ref{thm: iz_second_moment}. We examine both common-base and heterogeneous distribution families. In the common-base design of Equation~\eqref{eq: mean_shifting}, every alternative is a shifted copy of the same base random variable $X_0$, so it is enough to specify the distribution of $X_0$.
We consider three representative non-sub-Gaussian distributions for $X_0$: (1) the Lognormal distribution, parameterized by log-mean $\mu$ and log-standard deviation $\sigma$, with all moments being finite; (2) the Student's t distribution, parameterized by degree of freedom (df) and scale, which has finite $q$th moment only when $q$ is less than df; and (3) the Pareto distribution, parameterized by shape and scale, which has finite $q$th moment only when $q$ is smaller than the shape parameter.
This leads to six configurations: {SC-Lognormal}, {SC-Student's t}, {SC-Pareto}, {MM-Lognormal}, {MM-Student's t}, and {MM-Pareto}. The specific parameter values used in these configurations are reported in Section~\ref{subsec: sample_optimal_IZ}.
% We consider three representative distribution families: Lognormal, Student’s t, and Pareto. This gives rise to six benchmark configurations:
% {SC-Lognormal}, {SC-Student's t}, {SC-Pareto}, {MM-Lognormal}, {MM-Student's t}, and {MM-Pareto}. 
To evaluate performance without a common base distribution, we adopt the mixed-distribution configuration approach proposed by \citet{lee2016general}. We extend the mean-shifting model in Equation~\eqref{eq: mean_shifting} to involve two distinct base random variables, $X_{\text{odd}}$ and $X_{\text{even}}$, and define the alternative distributions as
\begin{equation}
\label{eq: mean_shifting_2}
    X_i \stackrel{d}{=} 
\begin{cases}
X_{\text{odd}}, & \text{if } i = 1, \\\\
X_{\text{odd}} - \gamma - \lambda (i/k)^{\beta}, & \text{if } i \geq 2 \text{ and } i \text{ is odd}, \\\\
X_{\text{even}} - \gamma - \lambda (i/k)^{\beta}, & \text{if } i \geq 2 \text{ and } i \text{ is even}.
\end{cases}
\end{equation}
By selecting different distributions for $X_{\text{odd}}$ and $X_{\text{even}}$, we construct problem instances that do not share a common base distribution. In particular, we consider the following two mixed-distribution configurations under both SC and MM configurations: (1) {Odd(t)-Even(Pareto)}, where $X_{\text{odd}}$ is Student's t and $X_{\text{even}}$ is Pareto; (2) {Odd(Pareto)-Even(t)}, where $X_{\text{odd}}$ is Pareto and $X_{\text{even}}$ is Student's t. Their parameters are selected to give finite, comparable variances, as required by Assumption~\ref{assu: bounded_variance}. The specific parameter values used in the experiments are listed in Section~\ref{subsec: sample_optimal_IZ}.
% We let $\gamma=0.1$. The parameters are set to let all alternatives have similar variances 0.4. The total sampling budget is 100$k$ for each $k$. Following \cite{lee2016general}, we also consider configurations with mixed distributions. (There is no a prior reason to believe that outputs from different alternatives follow the same distribution family). 

\subsubsection{Tested UCB Algorithms and Experimental Settings}
Under the IZ and uniformly bounded-variance assumptions, we have proved sample optimality for all UCB algorithms satisfying Assumption~\ref{assu: bonusfunction}. In our experiments, we evaluate several representative algorithms from the meta-UCB class: (1) {UCBE} \citep{audibert2010best}, where the UCB value is defined in Example \ref{example: ucbe};
    (2) {MOSS} \citep{audibert2010regret}, where the UCB value is defined in Example \ref{example: moss};
    (3) {Greedy}, which uses no bonus function (i.e., $f(n_i) = 0$) and always samples the alternative with the highest sample mean. 
    % This purely exploitative approach has been studied in \citet{li2023surprising} and shown to be sample optimal in large-scale Gaussian settings
Beyond this class, we also test the famous {UCB1} algorithm of \citet{auer2002finite}, which does not satisfy Assumption~\ref{assu: bonusfunction}. Its UCB value is defined as
\begin{eqnarray}
\label{eq: UCB1_index}
    U_i(n_i) = \bar X_i(n_i) + \sqrt{\frac{2 \log n}{n_i}},
\end{eqnarray}
where $n = \sum_{i=1}^k n_i$ is the overall sample size of all alternatives. Although {UCB1} is structurally similar to {UCBE} and {MOSS}, we will see that it may behave very differently due to its coupling across alternative through the global sample size $n$. Since our focus is not on algorithm tuning, we use fixed algorithm parameters (e.g., $a = 1$ for {UCBE}) across all experiments.

Unless otherwise specified, we evaluate algorithm performance across a range of problem sizes by varying the number of alternatives $k \in \{2^5, 2^6, \ldots, 2^{15}\}$. The total sampling budget is set as $B = c k$ for each $k$, where the constant $c$ may vary by configuration. For each algorithm, we estimate the PCS in solving any problem instance based on 500 independent macro replications.

\subsection{Sample Optimality under the Indifference-Zone Formulation}
\label{subsec: sample_optimal_IZ}
In this subsection, we illustrate the sample-optimal behavior predicted under the indifference-zone (IZ) formulation. For all SC and MM configurations, we set the IZ parameter $\gamma = 0.1$. In SC configurations, we let $\lambda = 0$, and the parameter $\beta$ is irrelevant. In MM configurations, we set $\lambda = 1$ and $\beta = 1$. To reflect different levels of problem difficulty, we use a total sampling budget of $B = 100k$ for each $k$ under SC configurations and $B = 30k$ for MM configurations. We choose the parameters of $X_0$, $X_{\text{odd}}$, and $X_{\text{even}}$ so that their variances are approximately 1. Thus all configurations satisfy Assumption~\ref{assu: bounded_variance}. The distributional information is summarized in Table~\ref{tab: distributional_info}.

\begin{table}[htbp]
\TABLE
 {Distributional Parameters and Available Moments of the Reference Random Variables\label{tab: distributional_info}}
{\begin{tabular}{ccccc}
\hline

\hline
{Experimental Design} &  {Random Variable} &  {Distribution Type} &  {Parameters} &  {$q$-th Moments Exist for} \\
\hline

\hline
\multirowcell{3}{Common-base \\ distribution family} 
& $X_0$     & Lognormal     & $ \mu = -2$, $\sigma = 1.45$ & $q <\infty$ \\
& $X_0$   & Student’s $t$ & df $= 3$, scale $= 0.6$               & $q <3$ \\
& $X_0$      & Pareto        & shape $= 3$, scale $= 1.2$            & $q <3$ \\
\hline

\hline
\multirowcell{2}{Heterogeneous \\ distribution families} 
& $X_{\text{odd}}$ or $X_{\text{even}}$    & Student’s $t$ & df $= 4$, scale $= 0.7$               & $q <4$ \\
& $X_{\text{odd}}$ or $X_{\text{even}}$     & Pareto        & shape $= 4$, scale $= 2.1$            & $q <4$ \\
\hline

\hline
\end{tabular}}{}
\end{table}

\subsubsection{Common-Base Distribution Family}
\label{subsubsec: common_base}
\begin{figure}[h]
         \FIGURE
    {\includegraphics[width=1\linewidth]{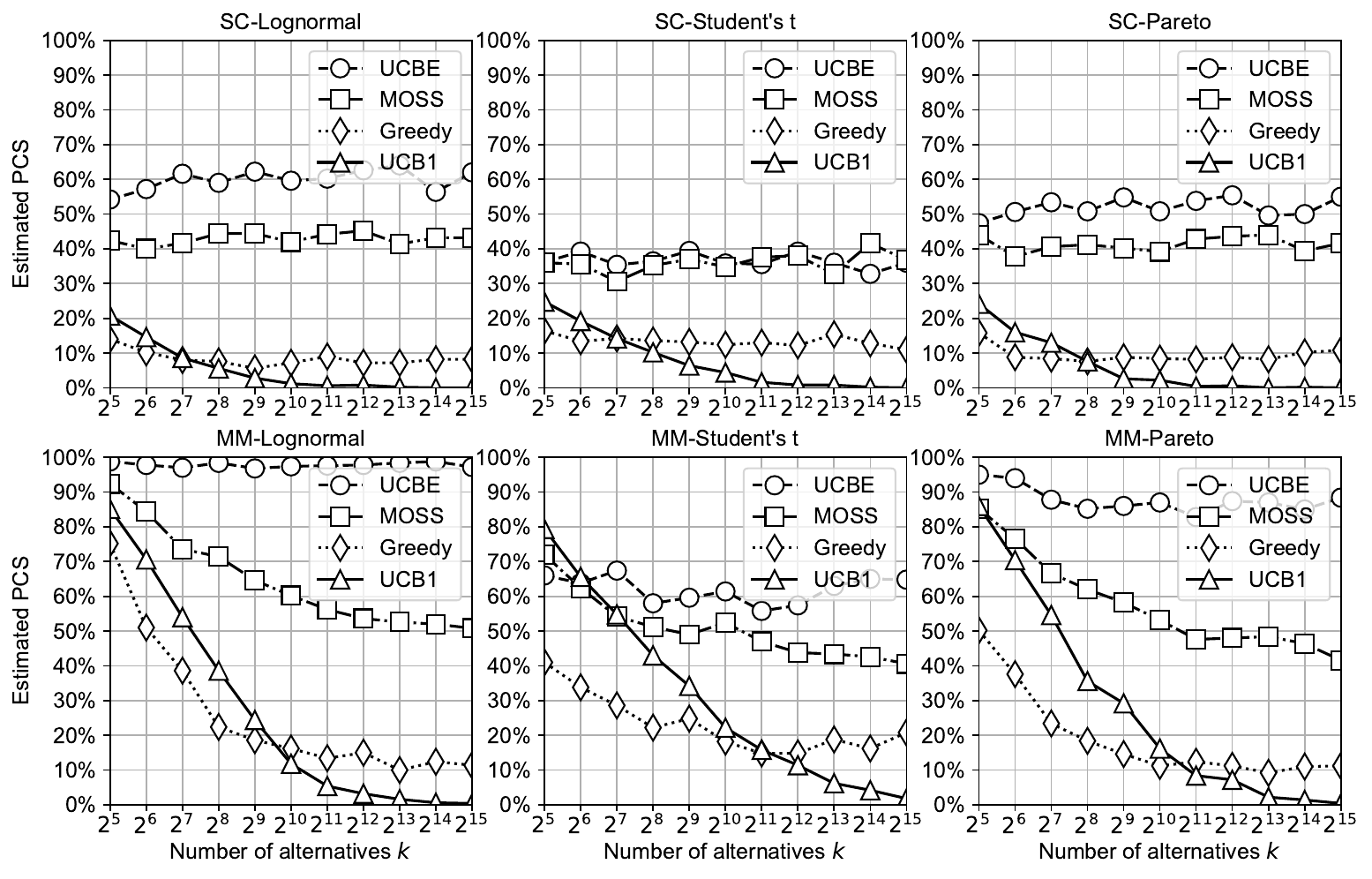}}
    {PCS of Different UCB Algorithms under Lognormal, Student's t, and Pareto Configurations.\label{fig:loc_scale_optimality}}{}
\end{figure}

 % with Lognormal, Student's t, and Pareto distributions
We first examine the performance of UCB algorithms when all alternatives are shifted copies of a common base distribution. Figure~\ref{fig:loc_scale_optimality} plots the PCS of each algorithm against $k$ under the SC and MM configurations. Several key observations can be made. First, under all configurations, the behavior of {UCBE}, {MOSS}, and {Greedy} is consistent with sample optimality. For instance, under the SC-Lognormal configuration, the PCS of {UCBE} remains above 50\% as $k$ increases. Although the PCS of {Greedy} is low when $k$ is large, it remains bounded away from zero over the displayed range. Second, both {UCBE} and {MOSS} significantly outperform {Greedy}, which has no bonus function. This illustrates the exploration effect of incorporating a bonus function satisfying Assumption \ref{assu: bonusfunction} in UCB algorithms. Interestingly, {UCBE} generally performs better than {MOSS}. This is possibly due to {MOSS}'s truncation structure, which caps the bonus and limits exploration for every alternative when its sample size exceeds $c$. It is also interesting to observe that the PCS levels may vary significantly across distribution families. In general, {UCBE} and {MOSS} perform better under Lognormal and Pareto configurations than under Student's t configurations. We explore this phenomenon further in the following experiment. Lastly, {UCB1} behaves quite differently: over the displayed range, its PCS steadily declines toward zero. We examine this behavior and its connection to {UCB1}’s structure in Section~\ref{subsec: budget_allocation}.

\subsubsection{Heterogeneous Distribution Families} 
\begin{figure}[htp]
         \FIGURE
    {\includegraphics[width=0.75\linewidth]{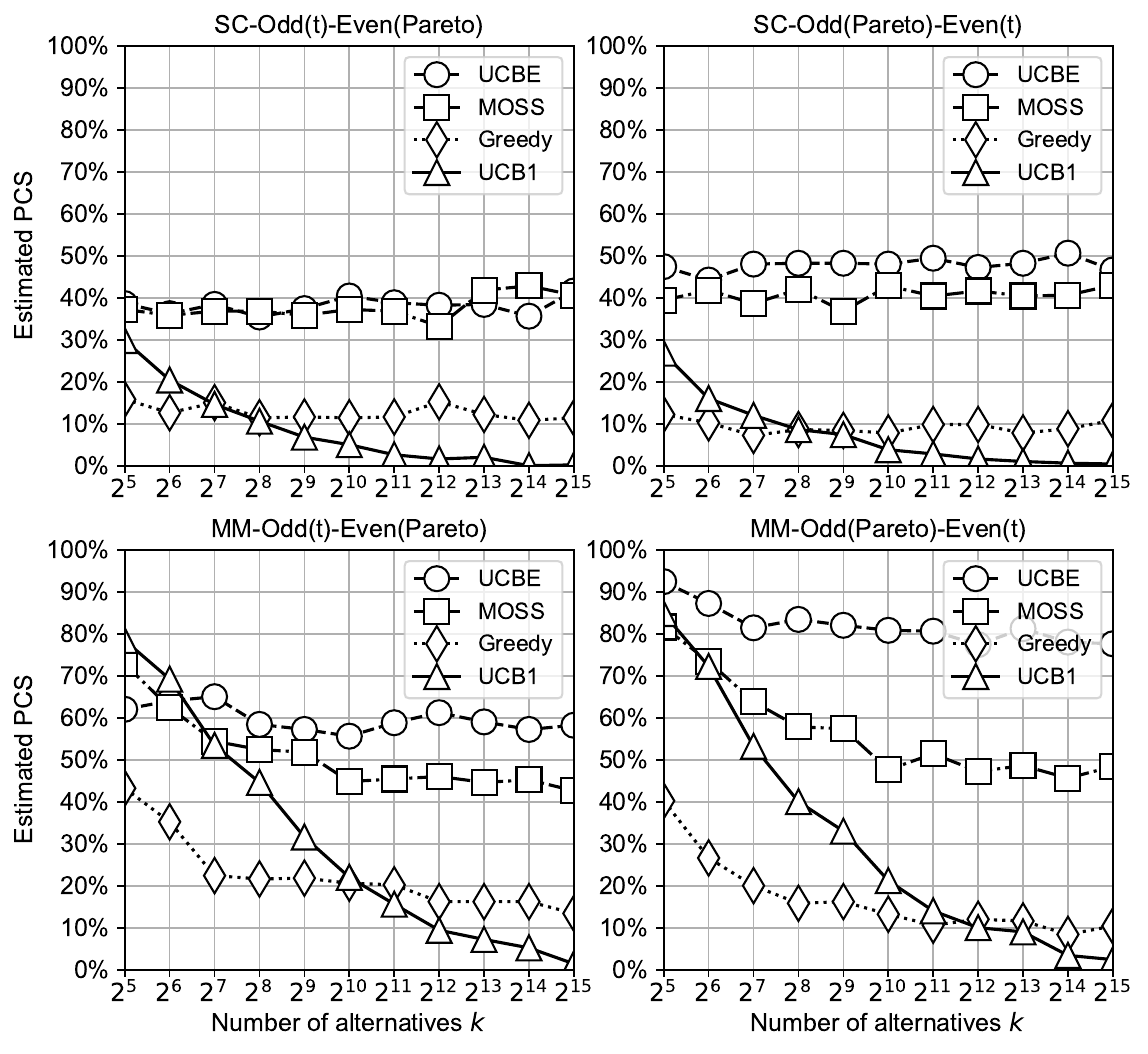}}
     {PCS under the Mixed-Distribution Configurations.\label{fig:moment_optimality}}{}

\end{figure}

We next examine mixed-distribution configurations that use different base distributions while retaining uniformly bounded variances. Figure~\ref{fig:moment_optimality} plots the PCS of each algorithm against $k$ under the SC and MM configurations. The findings are consistent with Figure~\ref{fig:loc_scale_optimality}. Most notably, the behavior of {UCBE}, {MOSS}, and {Greedy} remains consistent with sample optimality, whereas the PCS of {UCB1} again declines toward zero over the displayed range. We also observe that {UCBE} and {MOSS} perform better under the {Odd(P)-Even(t)} configurations than under the {Odd(t)-Even(Pareto)} configurations. This difference may be attributed to the distribution of the best alternative. Under the mean-shifting model in Equation~\eqref{eq: mean_shifting_2}, the key distinction is whether the best alternative follows a Student's t or a Pareto distribution. The Student's t distribution is symmetric with support over the entire real line, while the Pareto distribution is one-sided with support on $(\text{scale}, \infty)$. Observations from the Student's t distribution may therefore have a higher probability of taking very low values, making it harder for the best alternative to be re-selected after unfavorable initial observations, even with an exploration bonus. This insight may also help explain the trend in Figure~\ref{fig:loc_scale_optimality}, where algorithms tend to perform better under Lognormal and Pareto configurations.
% than under Student's t configurations.

 \subsection{ Budget Allocation Behaviors}
 \label{subsec: budget_allocation}
\begin{figure}[htp]
     \FIGURE
    {\includegraphics[width=1\linewidth]{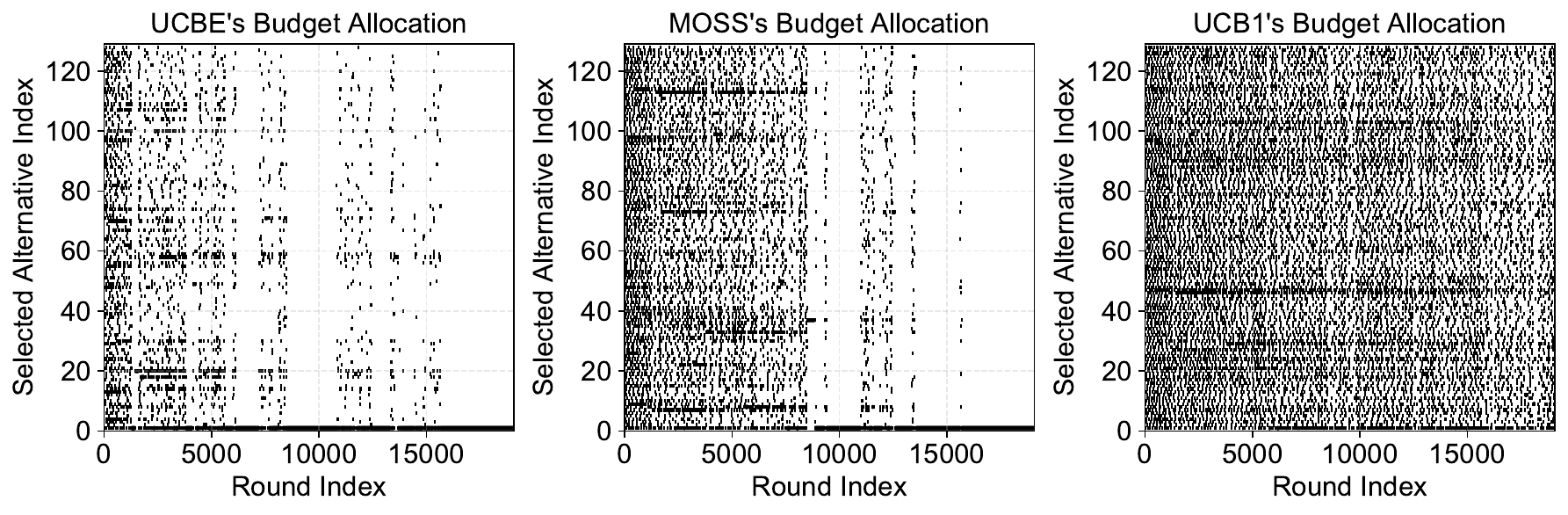}}
     {Budget Allocation in a Sample Path of Different UCB Algorithms for a Problem with $k=128$.     \label{fig:budget_allocation}}
    {Each round is marked by a dark dot indicating the alternative selected in that round.}

\end{figure}

\begin{figure}[htp]
     \FIGURE
    {\includegraphics[width=1\linewidth]{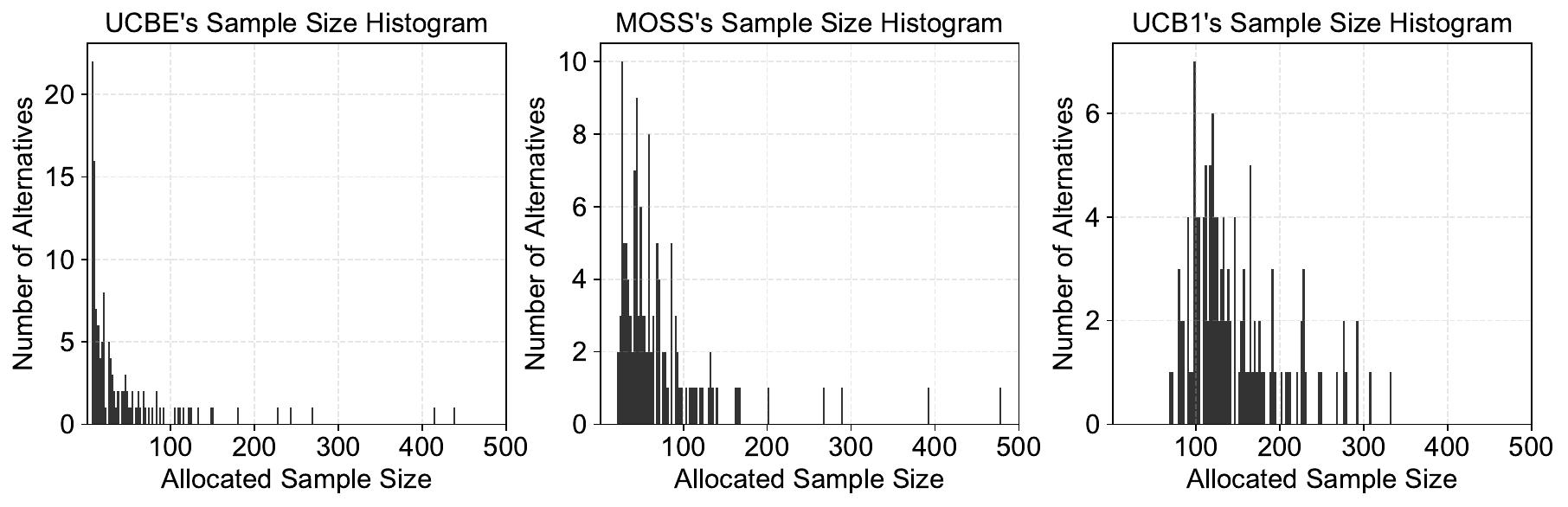}}
    {Allocated Sample Sizes of Inferior Alternatives for a Problem with $k=128$.  \label{fig:sample_size}}{}
  
\end{figure}

In this subsection, we compare the budget allocation behaviors of {UCBE}, {MOSS}, and {UCB1} to gain insight into their contrasting PCS behavior. To illustrate these allocation dynamics, we consider a relatively small-scale problem with \(k = 128\) alternatives under the {SC-Lognormal} configuration. We visualize the sample paths of all three algorithms under a common random seed and a total sampling budget of \(B = 150k\), where all algorithms correctly identify the best alternative within that sample path. These sample paths are shown in Figure~\ref{fig:budget_allocation}, where every sampling round is marked by a dark dot indicating the alternative selected in that round. To further illustrate the allocation behavior, we also plot the histogram of the allocated sample sizes of all \emph{inferior} alternatives ($i=2, \cdots, 128$) at the end of the sampling process in Figure~\ref{fig:sample_size}.

Figure~\ref{fig:budget_allocation} and Figure~\ref{fig:sample_size} show that {UCBE} and {MOSS} exhibit similar sample allocation behaviors, whereas {UCB1} behaves quite differently. For {UCBE} and {MOSS}, we observe the boundary-crossing dynamics described in Section~\ref{subsec: boundary-crossing}. After certain rounds, once the boundary-crossing processes of the inferior alternatives are completed, the algorithms focus almost exclusively on the best alternative. This transition may repeat several times. During the process that the UCB value of the best alternative drops to its minimum, some non-best alternatives may temporarily surpass it and get resampled to complete their boundary-crossing. Eventually, the boundary-crossing event described in Section~\ref{subsec: boundary-crossing} occurs, after which only the best alternative continues to be sampled. Furthermore, many inferior alternatives are sampled only a few times, as illustrated in the histogram in Figure~\ref{fig:sample_size}. We believe this selective sampling behavior is key to achieving sample optimality under suitable distributional conditions. In contrast, {UCB1} continues to allocate samples to most (if not all) inferior alternatives until the total sampling budget is exhausted, as shown in Figure~\ref{fig:budget_allocation}. This results in a more uniform allocation across alternatives, as illustrated in the histogram in Figure~\ref{fig:sample_size}. This behavior is somewhat surprising, as it reveals that {UCB1}—designed to strike an optimal balance between exploration \emph{and} exploitation in online learning—may actually engage in \emph{more} exploration than fixed-budget R\&S algorithms such as {UCBE} in large-scale problems.

This interesting behavior of {UCB1} appears to stem from the global sample size term $n=\sum_{i=1}^k n_i$ in the UCB value definition (Equation~\eqref{eq: UCB1_index}), which introduces coupling across alternatives. Because this term grows regardless of which alternative is selected, it causes the UCB value of all alternatives—including clearly inferior ones—to increase over time, potentially leading to unnecessary resampling. Such behavior may become particularly wasteful when the number of alternatives is large and may help explain the declining PCS of {UCB1} over the tested range. This contrast highlights the importance of the decoupling structure inherent in bonus functions that satisfy Assumption~\ref{assu: bonusfunction} for solving large-scale R\&S problems. 
% This highlights the importance of decoupling among alternatives for UCB algorithms in solving large-scale pure exploration problems—that is, the UCB value of each alternative should depend only on its own sample information.
% during some intermediate phases that the best alternative is continuously sampling, then turn to some inferior non-best alternatives; 

%  \begin{figure}[htp]
%     \centering
%     \includegraphics[width=1\linewidth]{Budget_Allocation.pdf}
%     \caption{Budget Allocation in a Sample Path of Different UCB Algorithms for a Problem with $k=128$. Each round is marked by a dark dot indicating the alternative selected in that round.}
%     \label{fig:budget_allocation}
% \end{figure}

\section{Concluding Remarks}
\label{sec: conclude}
In this paper, we explore the performance of UCB algorithms in large-scale R\&S problems under non-sub-Gaussian distributional assumptions. We introduce a meta-UCB algorithm with a general UCB value structure and develop a unified analysis framework based on a boundary-crossing perspective. This leads to a distribution-free lower bound on the PCS, which provides structural insights into the role of the exploration bonus function in balancing exploration and selection accuracy. Building on this PCS bound, we show that the meta-UCB algorithm achieves sample optimality under the indifference-zone (IZ) formulation and uniformly bounded variances. We establish sample optimality beyond the IZ formulation for polynomial gaps. We also conjecture that sample optimality can hold for a broader class of UCB algorithms satisfying a decoupling condition. Numerical experiments support the boundary-crossing perspective and further demonstrate the robustness of the UCB approach.

This paper opens several directions for future research. On the theoretical side, it is natural to ask whether either result can be extended below finite variance. 
% For the non-IZ formulation, it would also be useful to determine how far the conservative logarithmic inflation in Equation~\eqref{eq: log_sqrt_bonus} can be reduced under additional distributional structure or by replacing the raw sample mean with a robust center, and to understand the boundary case $\beta=1/2$.
% Establishing the sample optimality under weaker assumptions may require new, more robust UCB constructions. 
It is of interest to extend our results to a broader class of UCB algorithms with decoupled  UCB values, as discussed in Section \ref{subsec: more_general}, which are not covered by our current framework. Additionally, we have not discussed the issue of consistency, which requires that the PCS converges to 1 as the total sampling budget grows faster than the number of alternatives. This may be achieved by incorporating an explore-first framework \citep{li2023surprising} into UCB algorithms. Furthermore, from a practical perspective, the algorithm selection problem—i.e., identifying which UCB algorithms perform best under specific problem settings—remains an open question. Finally, as UCB algorithms are inherently sequential, developing parallel implementations for improved computational efficiency presents an important venue for future work.

\SingleSpacedXI
\bibliographystyle{informs2014}
\bibliography{ref.bib}
\OneAndAHalfSpacedXI
%%%%%%%%%%

%% Here starts the e-companion (EC)
%%%%%%%%%%%%%%%%%%%%%%%%%%%%%%%%%%%%%%%%%%%%%%%%%%%%%%%%%%
\ECSwitch

%\ECDisclaimer
%%%%%%%%%%%%%%%%%%%%%%%%%%%%%%%%%%%%%%%%%%%%%%%%%%%%%%%%%%

%%% Main head for the e-companion
\ECHead{E-companion to \\ \vspace{0.3cm}
{{Upper Confidence Bound Algorithms for Large-Scale Ranking and Selection: Beyond Sub-Gaussianity}}}
\section{Proofs}

\subsection{Proof of Lemma \ref{lem: linear_mean_subquadratic_variance}}
\label{subsec: proof_lem_linear_mean_subquadratic_variance}
\begin{proof}{Proof.}
For all sufficiently large $k$, the mean assumption gives
$\E[S_k]\leq Ck$. Hence, Chebyshev's inequality gives
\[
\Pr\left\{S_k\geq(C+\varepsilon)k\right\}
\leq
\Pr\left\{S_k-\E[S_k]\geq\varepsilon k\right\}
\leq
\frac{\operatorname{Var}(S_k)}{\varepsilon^2k^2}.
\]
Moreover, $\lim_{k\to\infty}\operatorname{Var}(S_k)/(\varepsilon^2k^2)=0$, which proves the result.
\hfill\Halmos
\end{proof}

We use the following standard ladder-epoch identity, which follows directly from Corollaries 8.39 and 8.44 of \cite{siegmund1985sequential}.
\begin{lemma}\label{lem: tau_-}
Let $X_1,X_2,\ldots$ be i.i.d. random variables. For a constant $b$, define $\bar{X}(n)=n^{-1}\sum_{i=1}^n X_i$ and $\tau=\inf\{n\geq1:\bar{X}(n)<b\}$. Then
\begin{eqnarray*}
    \mathrm{Pr}\{\tau=\infty\}= \exp\left(-\sum_{n=1}^\infty n^{-1}\mathrm{Pr}\{\bar{X}(n) < b\}\right) \quad \mbox{ and } \quad \mathrm{E}[\tau]=\exp\left(\sum_{n=1}^\infty n^{-1}\mathrm{Pr}\{\bar{X}(n) \geq b\}\right).
\end{eqnarray*}

\end{lemma}

\subsection{Proof of Lemma \ref{lem: PCS_bound}}
\label{subsec: proof_lem_pcs_bound}
\begin{proof}{Proof.}
By the definition of $U_1^*$,
\begin{eqnarray}
\label{eq: PCS_bounding_1}
 \Pr \left\{U^*_1 \geq \mu_1 - \gamma_0 \right\}  & = & \Pr \left\{\forall n \geq 1, \bar X_1(n) \geq \mu_1 - \gamma_0 - f(n) \right\} \geq \Pr \left\{\forall n \geq 1, \bar X_1(n) \geq \mu_1 - \gamma_0 \right\}, \qquad
\end{eqnarray}
where the inequality follows from Assumption~\ref{assu: bonusfunction}. Let $\tau=\inf\{n\geq1:\bar{X}_1(n)-\mu_1<-\gamma_0\}$. Then
\begin{eqnarray*}
    \Pr \left\{\forall n \geq 1, \bar X_1(n) \geq \mu_1 - \gamma_0 \right\} = \Pr \left\{\tau=\infty\right\}. 
\end{eqnarray*}
For $\gamma_0>0$, Lemma~\ref{lem: tau_-} gives
\begin{eqnarray}
\label{eq: PCS_bounding_2}
\Pr \left\{\forall n \geq 1, \bar X_1(n) \geq \mu_1 - \gamma_0 \right\}
&=& \Pr \left\{\tau=\infty\right\} \nonumber= \exp \left(-\sum_{n=1}^{\infty} \frac{1}{n} \Pr\left\{ \bar{X}_1(n)-\mu_1< -\gamma_0 \right\}\right) \nonumber \\
&\geq& \exp \left(-\sum_{n=1}^{\infty} \frac{1}{n} \Pr\left\{|\bar{X}_1(n) - \mu_1|\geq \gamma_0 \right\}\right) \nonumber \\
% &\geq& \exp \left(-\sum_{n=1}^{\infty} \frac{1}{n} \frac{\mathrm{Var}\left[ \bar{X}_1(n)-\mu_1 \right]}{\gamma_0^2}\right) \qquad \mbox{(by Chebyshev's inequality)} \nonumber \\
&\geq& \exp \left(-\sum_{n=1}^{\infty} \frac{\sigma_1^2}{n^2 \gamma_0^2}\right) \qquad \mbox{(by Chebyshev's inequality)} \nonumber \\
&=& \exp \left(-\frac{\pi^2 \sigma_1^2}{6\gamma_0^2}\right),
\end{eqnarray}
which is strictly positive. Combining Equations~\eqref{eq: PCS_bounding_1} and~\eqref{eq: PCS_bounding_2} proves the claim. \hfill\Halmos
\end{proof}

\subsection{Proof of Lemma \ref{lem: move}}
\label{subsec: proof_lem_move}
\begin{proof}{Proof.}
If $\bar T^f(b/2)=\infty$, the claimed inequality is immediate. Otherwise, let $n=\bar T^f(b/2)$. Since $n\geq n^f(b/2)$, the definition of $n^f(b/2)$ gives $f(n)\leq b/2$. The definition of $\bar T^f(b/2)$ also gives $\bar X(n)<b/2$. Hence
\[
\bar X(n)<\frac b2\leq b-f(n),
\]
so $n$ is a valid crossing time for the process on the left-hand side. \hfill\Halmos
\end{proof}

\subsection{Proof of Lemma \ref{lem: finite_variance_crossing}}
\label{subsec: proof_lem_finite_variance_crossing}
\begin{proof}{Proof.}
Define the nonoverlapping block averages
\[
W_r=\frac1{n_0}\sum_{j=(r-1)n_0+1}^{rn_0}Y_j,
\qquad r\geq1.
\]
They are i.i.d. with $\E[W_r]=0$ and $v:=\operatorname{Var}(W_r)=\sigma^2/n_0$.
Let $M_0=R_0=0$. For $m\geq1$, set $M_m=\sum_{r=1}^mW_r$, $R_m=M_m-bm$, and $\tau=\inf\{m\geq1:R_m<0\}$.
On $\{\tau<\infty\}$, the average of the first $n_0\tau$ original
observations is $M_\tau/\tau<b$; on $\{\tau=\infty\}$, the following
inequality is automatic. Therefore,
\begin{equation}
\label{eq: delayed_crossing_block_bound}
H(b;n_0)\leq n_0\tau.
\end{equation}

By Lemma~\ref{lem: tau_-},
\[
\E[\tau]
=\exp\left\{\sum_{m=1}^{\infty}\frac1m
\Pr(M_m\geq bm)\right\}.
\]
Chebyshev's inequality gives
\[
\Pr(M_m\geq bm)
\leq\frac{\operatorname{Var}(M_m)}{m^2b^2}
=\frac{v}{mb^2}.
\]
Substituting this bound into the ladder-epoch identity and using
$\sum_{m\geq1}m^{-2}=\pi^2/6$ yields
\begin{equation}
\label{eq: ladder_epoch_mean_bound}
\E[\tau]
\leq
\exp\left(\frac{v}{b^2}\sum_{m=1}^{\infty}\frac1{m^2}\right)
=\exp\left(\frac{\pi^2v}{6b^2}\right)<\infty.
\end{equation}
In particular, $\tau<\infty$ almost surely.

We now control the second moment. Let $\mathcal F_m=\sigma(W_1,\ldots,W_m)$.
Since $\E[W_{m+1}\mid\mathcal F_m]=0$ and
$\E[W_{m+1}^2\mid\mathcal F_m]=v$, we have
$\E[M_{m+1}^2\mid\mathcal F_m]=M_m^2+v$. Thus $M_m^2-mv$ is a
martingale. For $L\geq1$, the truncated crossing time
$\tau_L=\tau\wedge L$ is a bounded stopping time, so the optional stopping
theorem gives
\[
\E[M_{\tau_L}^2-\tau_Lv]=\E[M_0^2]=0.
\]
Therefore, $\E[M_{\tau_L}^2]=v\E[\tau_L]$.
As $L\to\infty$, $M_{\tau_L}\to M_\tau$ almost surely. Fatou's lemma and monotone convergence therefore yield
\begin{equation}
\label{eq: stopped_sum_second_moment}
\E[M_\tau^2]
\leq\liminf_{L\to\infty}\E[M_{\tau_L}^2]
=v\lim_{L\to\infty}\E[\tau_L]
=v\E[\tau].
\end{equation}

It remains to control the overshoot. On $\{\tau=m\}$, we have $R_{m-1}\geq0$ and $R_m<0$, and hence
\[
0<-R_m=b-W_m-R_{m-1}\leq(b-W_m)^+.
\]
Since $R_\tau^2=\sum_{m\geq1}R_m^2\mathbf 1\{\tau=m\}$, this implies
\[
R_\tau^2
\leq\sum_{m=1}^{\infty}(b-W_m)_+^2\mathbf 1\{\tau=m\}
\leq\sum_{m=1}^{\infty}(b-W_m)_+^2\mathbf 1\{\tau\geq m\}.
\]
All summands are nonnegative, so Tonelli's theorem gives
\begin{eqnarray}
\label{eq: predictable_overshoot_bound}
\E[R_\tau^2]
&\leq&\sum_{m=1}^{\infty}
\E\left[(b-W_m)_+^2\mathbf 1\{\tau\geq m\}\right]\nonumber
=\E[(b-W_1)_+^2]
\sum_{m=1}^{\infty}\Pr(\tau\geq m)\nonumber
=\E[(b-W_1)_+^2]\E[\tau]\nonumber
 \leq (b^2+v)\E[\tau].
\end{eqnarray}
Indeed, $\{\tau\geq m\}=\{R_1\geq0,\ldots,R_{m-1}\geq0\}$ belongs to
$\mathcal F_{m-1}$ and is therefore independent of $W_m$. The second
equality uses the tail-sum identity
$\sum_{m\geq1}\Pr(\tau\geq m)=\E[\tau]$, while the last inequality follows
from $\E[(b-W_1)_+^2]\leq\E[(b-W_1)^2]=b^2+v$. This argument uses only
the independence of $W_m$ from the past; it does not require $W_\tau$ to
be independent of $\tau$.

Finally, $b\tau=M_\tau-R_\tau$. Using $(x-y)^2\leq2x^2+2y^2$ together
with Equations~\eqref{eq: stopped_sum_second_moment}
and~\eqref{eq: predictable_overshoot_bound} gives
\[
b^2\E[\tau^2]
\leq2\E[M_\tau^2]+2\E[R_\tau^2]
\leq(4v+2b^2)\E[\tau].
\]
Equation~\eqref{eq: delayed_crossing_block_bound} gives
$\E[H(b;n_0)]\leq n_0\E[\tau]$ and
$\E[H(b;n_0)^2]\leq n_0^2\E[\tau^2]$. Substituting the preceding
bounds and $v=\sigma^2/n_0$ proves both inequalities in
Equation~\eqref{eq: finite_variance_crossing_bounds}. \hfill\Halmos
\end{proof}

\subsection{Proof of Theorem \ref{thm: iz_second_moment}}
\label{subsec: proof_thm_iz_second_moment}
\begin{proof}{Proof.}
Fix $\gamma_0\in(0,\gamma)$ and let $S_k=\sum_{i=2}^kT_i^f(\mu_1-\gamma_0)$. These crossing times are independent because they depend on mutually independent observation streams.
Moreover,
$\operatorname{Var}(S_k)=\sum_{i=2}^k
\operatorname{Var}(T_i^f(\mu_1-\gamma_0))$, and each variance is at most
the corresponding second moment. Thus, Equation~\eqref{eq: iz_crossing_constants} gives
\[
\E[S_k]\leq(k-1)C_2^f(\gamma_0),
\qquad
\operatorname{Var}(S_k)\leq(k-1)D_2^f(\gamma_0).
\]
Consequently, Lemma~\ref{lem: linear_mean_subquadratic_variance} gives, for every $\varepsilon>0$,
\begin{equation}
\label{eq: iz_rowwise_chebyshev}
\lim_{k\to\infty}
\Pr\left\{S_k\geq\bigl(C_2^f(\gamma_0)+\varepsilon\bigr)k\right\}
=0.
\end{equation}

Let $\mathcal A_k=\{U_1^*\geq\mu_1-\gamma_0\}$. On $\mathcal A_k$, monotonicity of the boundary-crossing times gives
\[
T_i^f(U_1^*)
\leq T_i^f(\mu_1-\gamma_0),
\qquad i=2,\ldots,k.
\]
Choose $\varepsilon\in(0,c/2-C_2^f(\gamma_0))$. Since $B=ck$, Equation~\eqref{eq: iz_rowwise_chebyshev} implies
\[
\lim_{k\to\infty}\Pr\{B>2S_k\}=1.
\]
Indeed, $2(C_2^f(\gamma_0)+\varepsilon)k<ck=B$, so
$\{B\leq2S_k\}$ is contained in the event whose probability tends to
zero in Equation~\eqref{eq: iz_rowwise_chebyshev}.
On $\mathcal A_k\cap\{B>2S_k\}$, the preceding comparison gives
$B>2\sum_{i=2}^kT_i^f(U_1^*)$. Hence Equation~\eqref{eq: PCS_bound}
implies
\[
\mathrm{PCS}\geq\Pr\bigl(\mathcal A_k\cap\{B>2S_k\}\bigr).
\]
The event $\mathcal A_k$ depends only on the best-alternative stream,
whereas $S_k$ depends only on the non-best streams, so the two events
are independent. Lemma~\ref{lem: PCS_bound} and
$\sigma_1^2\leq\bar\sigma^2$ also give
$\Pr(\mathcal A_k)\geq
\exp\{-\pi^2\bar\sigma^2/(6\gamma_0^2)\}$ uniformly in $k$.
Therefore,
\begin{eqnarray*}
\liminf_{k\to\infty}\mathrm{PCS}
&\geq&\liminf_{k\to\infty}
\Pr(\mathcal A_k)\Pr\{B>2S_k\} \geq \exp\left(-\frac{\pi^2\bar\sigma^2}{6\gamma_0^2}\right)>0.
\end{eqnarray*}
Taking any $\alpha$ strictly below the final display verifies Definition~\ref{def: rate_optimality_PCS}. \hfill\Halmos
\end{proof}

\subsection{Proof of Lemma \ref{lem: variance_CS}}
\label{subsec: proof_lem_variance_CS}
\begin{proof}{Proof.}
Let $S_n=\sum_{\ell=1}^n(X_{1,\ell}-\mu_1)$.
For $j\geq0$ and $2^j\leq n<2^{j+1}$,
\[
nf_A(n)
=A\sqrt n\,\log_2(2n)
\geq A2^{j/2}(j+1).
\]
If $\bar X_1(n)+f_A(n)<\mu_1$ for some $n$ in this epoch, then
$S_n<-nf_A(n)$, and hence
$\max_{1\leq m\leq2^{j+1}}|S_m|\geq A2^{j/2}(j+1)$.
The increments of $S_m$ are independent and centered, with
$\operatorname{Var}(S_{2^{j+1}})=2^{j+1}\sigma_1^2$.
Kolmogorov's maximal inequality therefore gives
\begin{eqnarray*}
&&\Pr\left\{\exists n\in[2^j,2^{j+1}):
\bar X_1(n)+f_A(n)<\mu_1\right\}
\leq 
\Pr\left\{\max_{1\leq m\leq2^{j+1}}|S_m|
\geq A2^{j/2}(j+1)\right\}
\leq 
\frac{2^{j+1}\sigma_1^2}
{A^2 2^j(j+1)^2}
=\frac{2\sigma_1^2}{A^2(j+1)^2}.
\end{eqnarray*}
A union bound over the dyadic epochs yields
\[
\Pr\left\{\exists n\geq1:
\bar X_1(n)+f_A(n)<\mu_1\right\}
\leq
\frac{2\sigma_1^2}{A^2}
\sum_{j=0}^{\infty}\frac1{(j+1)^2}
=\frac{\pi^2\sigma_1^2}{3A^2}
\leq\alpha.
\]
Finally, $f_A$ is nonnegative and converges to zero, so it satisfies
Assumption~\ref{assu: bonusfunction}. \hfill\Halmos
\end{proof}

\subsection{Proof of Lemma \ref{lem: log_bonus_inverse}}
\label{subsec: proof_lem_log_bonus_inverse}
\begin{proof}{Proof.}
For $b>0$, define
\[
N_A(b)=\left\lceil
\max\left\{2,\frac{16A}{b}\log\left(e+\frac{A}{b}\right)\right\}^{\!2}
\right\rceil.
\]
Write $r=A/b$, $\ell=\log(e+r)$, and $x=\max\{2,16r\ell\}$.
The function $t\mapsto\log_2(2t)/\sqrt t$ is decreasing for
$t\geq4$, because its derivative has the sign of $2-\log(2t)$.
If $x=2$, then $r\ell\leq1/8$, so $r\leq1/8$ and
\[
\frac{f_A(x^2)}b=\frac32r\leq\frac3{16}<1.
\]
If $x=16r\ell$, then
\begin{eqnarray*}
\frac{f_A(x^2)}b
&=&\frac{\log(2x^2)}{16\log(2)\ell}
=\frac{\log2+2\log16+2\log r+2\log\ell}
{16\log(2)\ell}
\leq\frac{9\log2+4}{16\log2}<1,
\end{eqnarray*}
where $\ell\geq1$, $\log r\leq\ell$, $\log\ell\leq\ell$, and
$4<7\log2$. Since $N_A(b)=\lceil x^2\rceil\geq x^2\geq4$,
monotonicity gives $f_A(n)\leq b$ for every $n\geq N_A(b)$.

For $0<b\leq1/2$, let $L_b=1+\log(1/b)$. Because
\[
\log\left(e+\frac Ab\right)
\leq\log(e+A)+\log(1/b)
\leq\log(e+A)L_b,
\]
we obtain
\begin{eqnarray*}
N_A(b)
&\leq&
5+256A^2b^{-2}\log^2(e+A)L_b^2 \leq 
\left[1+256A^2\log^2(e+A)\right]b^{-2}L_b^2.
\end{eqnarray*}
The last inequality uses
$b^{-2}L_b^2\geq4(1+\log2)^2>5$.
Thus, the upper bound in Lemma~\ref{lem: log_bonus_inverse} holds with
$\kappa_A=1+256A^2\log^2(e+A)$.
Finally, the definition of $N_A(b)$ directly gives
\[
N_A(b)b^2
\geq256A^2\log^2(e+A/b)
\geq256A^2.
\]
Hence $\inf_{b>0}b^2N_A(b)\geq256A^2>0$.
\hfill\Halmos
\end{proof}

\subsection{Proof of Lemma \ref{lem: noniz_second_moment_cost}}
\label{subsec: proof_lem_noniz_second_moment_cost}
\begin{proof}{Proof.}
For brevity, write $N_{i,k}=N_A(b_i)$ and $b_i=\Delta_i/2$.
Let $\rho_A=\inf_{b>0}b^2N_A(b)>0$. Applying
Lemma~\ref{lem: finite_variance_crossing} with delayed start
$N_{i,k}$ and boundary $b_i$ therefore yields the following bounds;
here $Y_j=X_{i,j}-\mu_i$, $n_0=N_{i,k}$, $b=b_i$, and
$\sigma^2=\sigma_i^2$ in that lemma.
\begin{eqnarray}
\label{eq: noniz_individual_crossing_bounds}
\E[\widehat T_{i,k}]
&\leq&\Gamma_A N_{i,k},\nonumber\\
\E[\widehat T_{i,k}^2]
&\leq&\Lambda_A N_{i,k}^2,
\end{eqnarray}
where
\[
\Gamma_A=
\exp\left(\frac{\pi^2\bar\sigma^2}{6\rho_A}\right),
\qquad
\Lambda_A=
\left(2+\frac{4\bar\sigma^2}{\rho_A}\right)\Gamma_A.
\]
Indeed, $b_i^2N_{i,k}\geq\rho_A$ and
$\operatorname{Var}(X_i)\leq\bar\sigma^2$, so the exponential and
prefactor in Lemma~\ref{lem: finite_variance_crossing} are bounded by
$\Gamma_A$ and $\Lambda_A$, respectively.

We first aggregate the means. Since $0<\Delta_i\leq1$, we have
$0<b_i\leq1/2$. Therefore, for $\Delta_i=(i/k)^\beta$,
Lemma~\ref{lem: log_bonus_inverse}
implies, with $a=1+\log2$,
\begin{eqnarray}
\label{eq: noniz_delayed_start_bound}
N_{i,k}
&\leq&
4\kappa_A
\left(\frac{k}{i}\right)^{2\beta}
\left[a+\beta\log(k/i)\right]^2,
\qquad a=1+\log2.
\end{eqnarray}
Here we used $b_i^{-2}=4(k/i)^{2\beta}$ and
$1+\log(1/b_i)=a+\beta\log(k/i)$.
The function $g(x)=x^{-2\beta}[a+\beta\log(1/x)]^2$ is decreasing and integrable on $(0,1]$ because $2\beta<1$.
Let
\[
J_\beta=
\int_0^1x^{-2\beta}\left[a+\beta\log(1/x)\right]^2\,dx
=
\frac{a^2}{1-2\beta}
+\frac{2a\beta}{(1-2\beta)^2}
+\frac{2\beta^2}{(1-2\beta)^3}.
\]
Because $g$ is decreasing,
$g(i/k)\leq k\int_{(i-1)/k}^{i/k}g(x)\,dx$ for
$i=2,\ldots,k$. Therefore,
\begin{eqnarray*}
\sum_{i=2}^kN_{i,k}
&\leq&
4\kappa_Ak\int_0^1
x^{-2\beta}\left[a+\beta\log(1/x)\right]^2\,dx =4\kappa_AJ_\beta k.
\end{eqnarray*}

The displayed expression for $J_\beta$ uses the identity
\[
\int_0^1x^{-p}\log^m(1/x)\,dx
=\frac{m!}{(1-p)^{m+1}},
\qquad p<1.
\]
Thus, the first assertion of Lemma~\ref{lem: noniz_second_moment_cost}
follows from Equation~\eqref{eq: noniz_individual_crossing_bounds} with $D_{\beta,A}=4\kappa_A\Gamma_AJ_\beta$.

It remains to aggregate the second moments. Put $\delta=1-2\beta>0$ and $R_\beta=8a^4+8\beta^4(4/(e\delta))^4$.
For every $t\geq0$,
\[
(a+\beta t)^4
\leq8a^4+8\beta^4t^4
\leq R_\beta e^{\delta t},
\]
because
$t^4\leq(4/(e\delta))^4e^{\delta t}$.
Taking $t=\log(k/i)$ gives $e^{\delta t}=(k/i)^\delta$ and
$4\beta+\delta=1+2\beta$. Hence
Equation~\eqref{eq: noniz_delayed_start_bound} gives
\begin{eqnarray*}
\sum_{i=2}^kN_{i,k}^2
&\leq&
16\kappa_A^2
\sum_{i=2}^k
\left(\frac{k}{i}\right)^{4\beta}
\left[a+\beta\log(k/i)\right]^4 \leq 
16\kappa_A^2R_\beta
\sum_{i=2}^k
\left(\frac{k}{i}\right)^{1+2\beta} \leq 
16\kappa_A^2R_\beta
\zeta(1+2\beta)k^{1+2\beta}
=o(k^2).
\end{eqnarray*}
The last inequality uses
$\sum_{i=2}^ki^{-(1+2\beta)}\leq\zeta(1+2\beta)$, and the final
relation follows from $1+2\beta<2$.
Combining this display with the second inequality in
Equation~\eqref{eq: noniz_individual_crossing_bounds} proves the
second assertion of Lemma~\ref{lem: noniz_second_moment_cost}.
\hfill\Halmos
\end{proof}

\subsection{Proof of Proposition \ref{prop: non-iz-optimality}}
\label{subsec: proof_prop_noniz_second_moment}
\begin{proof}{Proof.}
Let $\widehat S_k=\sum_{i=2}^k\widehat T_{i,k}$.
Because the non-best alternatives have mutually independent observation
streams,
$\operatorname{Var}(\widehat S_k)=
\sum_{i=2}^k\operatorname{Var}(\widehat T_{i,k})$.
Since $\operatorname{Var}(Z)\leq\E[Z^2]$, Lemma~\ref{lem: noniz_second_moment_cost} gives
\[
\E[\widehat S_k]\leq D_{\beta,A}k,
\qquad
\operatorname{Var}(\widehat S_k)
\leq\sum_{i=2}^k\E[\widehat T_{i,k}^2]
=o(k^2).
\]
Thus, Lemma~\ref{lem: linear_mean_subquadratic_variance} yields, for every $\varepsilon>0$,
\begin{equation}
\label{eq: noniz_rowwise_chebyshev}
\lim_{k\to\infty}
\Pr\left\{\widehat S_k\geq
(D_{\beta,A}+\varepsilon)k\right\}
=0.
\end{equation}

Let $\mathcal E_k=\{U_1^*\geq\mu_1\}$.
Lemma~\ref{lem: variance_CS} gives
$\Pr(\mathcal E_k)\geq1-\alpha$ uniformly in $k$. On
$\mathcal E_k$, monotonicity of the boundary-crossing times and
Equation~\eqref{eq: bc_f} give
\[
T_i^{f_A}(U_1^*)
\leq T_i^{f_A}(\mu_1)
\leq\widehat T_{i,k},
\qquad i=2,\ldots,k.
\]
Choose $\varepsilon\in(0,c/2-D_{\beta,A})$. Since
$B=ck$, Equation~\eqref{eq: noniz_rowwise_chebyshev}
implies $\lim_{k\to\infty}\Pr\{B>2\widehat S_k\}=1$.
Indeed, $B/2>(D_{\beta,A}+\varepsilon)k$, so
$\{B\leq2\widehat S_k\}$ is contained in the event whose probability
tends to zero in Equation~\eqref{eq: noniz_rowwise_chebyshev}.
On $\mathcal E_k\cap\{B>2\widehat S_k\}$, the preceding inequalities give
$B>2\sum_{i=2}^kT_i^{f_A}(U_1^*)$. Thus Equation~\eqref{eq: PCS_bound}
implies
\[
\mathrm{PCS}\geq
\Pr\bigl(\mathcal E_k\cap\{B>2\widehat S_k\}\bigr).
\]
The event $\mathcal E_k$ depends only on the best-alternative stream,
whereas $\widehat S_k$ depends only on the non-best streams, so the two
events are independent. Therefore,
\begin{eqnarray*}
\liminf_{k\to\infty}\mathrm{PCS}
&\geq&
\liminf_{k\to\infty}
\Pr(\mathcal E_k)\Pr\{B>2\widehat S_k\}\\
&\geq&1-\alpha>0.
\end{eqnarray*}
This proves sample optimality. \hfill\Halmos
\end{proof}

\end{document}